\documentclass[twoside,11pt]{article}

\usepackage{jmlr2e}
\usepackage {hyperref}
\usepackage{dirtytalk}
\usepackage{tikz}
\usepackage{pgfplots}
\usepackage{multirow}
\usepackage{subfigure}
\usepackage{amsmath}
\usepackage{booktabs}
\usepackage{xcolor}

\usepackage{xcolor}

\jmlrheading{1}{2024}{1-39}{11/01}{11/01}{Anvitha Thirthapura Sreedhara}

\begin{document}

% \title{Benchmark for time series forecasting - Sktime vs Autogluon-Timeseries}

\begin{center}
     \Large \textbf{CAN TIME SERIES FORECASTING BE AUTOMATED? A BENCHMARK AND ANALYSIS}
\end{center}
\vspace{0.075cm}
\begin{flushleft}   
    \textbf{Anvitha Thirthapura Sreedhara} \\
     {\scriptsize\MakeUppercase{anvithats20@gmail.com}} \\
    \textit{Department of Mathematics and Computer Science} \\
    \textit{Technical University of Eindhoven}
\end{flushleft}

\begin{flushright}   
    \textbf{Joaquin Vanschoren} \\
     {\scriptsize\MakeUppercase{j.vanschoren@tue.nl}} \\
    \textit{Department of Mathematics and Computer Science} \\
    \textit{Technical University of Eindhoven}
\end{flushright}
% \title{CAN TIME SERIES FORECASTING BE AUTOMATED? A BENCHMARK AND ANALYSIS}
% \author{\name Anvitha Thirthapura Sreedhara \\
% \email anvithats20@gmail.com \\
%        \addr Department of Mathematics and Computer Science\\
%        Technical University of Eindhoven}

%\editor{Anvitha Thirthapura Sreedhara} %ask professor
%\maketitle

%a.thirthapura.sreedhara@student.tue.nl
\vspace{1.2cm}

\begin{abstract}%   <- trailing '%' for backward compatibility of .sty file

In the field of machine learning and artificial intelligence, time series forecasting plays a pivotal role across various domains such as finance, healthcare, and weather. However, the task of selecting the most suitable forecasting method for a given dataset is a complex task due to the diversity of data patterns and characteristics. This research aims to address this challenge by proposing a comprehensive benchmark for evaluating and ranking time series forecasting methods across a wide range of datasets. This study investigates the comparative performance of many methods from two prominent time series forecasting frameworks, AutoGluon-Timeseries, and sktime to shed light on their applicability in different real-world scenarios. This research contributes to the field of time series forecasting by providing a robust benchmarking methodology, and facilitating informed decision-making when choosing forecasting methods for achieving optimal prediction. \\

%The results of this study will be valuable for practitioners and researchers seeking to make data-driven decisions in time series forecasting applications.

\end{abstract}


\section{Introduction}
\label{sec:introduction}

In recent times, the explosion of data has been remarkable, and the demand for machine-learning solutions has surged. Time series data, which represents a sequence of data points ordered by time, has gained special prominence due to its ability to capture temporal dependencies, trends, and patterns. Various industries, including retail and healthcare, rely on time series data to optimize inventory management, monitor patient information, and make informed decisions. Time series forecasting is a fundamental task with a wide range of applications, from supply chain management to weather forecasting and stock market analysis. However, the choice of an appropriate forecasting method is crucial, as it directly influences the quality and reliability of predictions.

Furthermore, the process of constructing an end-to-end machine-learning pipeline can be time-consuming. This has led to the emergence of automated machine learning (AutoML) tools designed to streamline and automate the process. Several AutoML tools, such as \href{https://github.com/winedarksea/AutoTS}{\textit{AutoTS}}, GAMA \citep{gijsbers2021gama}, AutoGluon-Timeseries \citep{shchur2023autogluon}, Auto-sklearn \citep{feurer-arxiv20a},  MLJAR \citep{mljar}, TPOT \citep{OlsonGECCO2016},  and others, have been developed, some of which are specifically tailored for time series forecasting. Nevertheless, the exploration of AutoML in the field of time series forecasting is still evolving and not yet widely explored.

Despite significant advancements in the field in recent years, it remains challenging to identify the most suitable technique for a given time series dataset, especially when datasets exhibit diverse temporal characteristics such as seasonality, trends, stationarity, etc. Hence, this research attempts to address these challenges and provide a systematic benchmark to assess different methods and gain insights on when and how to apply them. To achieve this, we propose the following research questions:

\begin{itemize}
    \item \textbf{RQ 1:} How can we develop a comprehensive benchmark for the evaluation of time series forecasting methods across a diverse range of datasets?  

    \item \textbf{RQ 2:} What is the comparative performance of automl methods and various individual forecasting algorithms from sktime in time series forecasting, and in what scenarios does AutoGluon excel over sktime, or vice versa?  

    \item \textbf{RQ 3:} How much can tuning help in the performance of sktime methods, specifically whether tuned sktime algorithms can rival the best automl methods? 
    
\end{itemize}

\subsection*{\textbf{RQ 1}}

A benchmark is designed, encompassing a substantial collection of time series datasets that have been carefully curated to represent a wide range of temporal characteristics. These datasets are sourced from the \textbf{Monash Time Series Forecasting Repository} \citep{godahewa2021monash} to ensure a holistic assessment of forecasting methods. The datasets are heterogeneous and cover varied domains including tourism, banking, energy, economics,  transportation, nature, web,  sales, and health. To evaluate the performance of forecasting frameworks on these varied datasets, different evaluation metrics are defined (Details in Section \ref{sec:evaluationmetrics}).

\subsection*{\textbf{RQ 2}}

A comprehensive evaluation is conducted on the AutoGluon-Timeseries (Section \ref{sec:autogluon}) and sktime (Section \ref{sec:sktime}) frameworks. This involves a thorough analysis to unveil valuable insights into the strengths and weaknesses of various time series forecasting techniques. These findings will guide practitioners seeking to make well-informed decisions when selecting forecasting methods for specific applications.

\subsection*{\textbf{RQ 3}}

The potential improvements in the performance of sktime methods through tuning and whether these tuned sktime algorithms can compete with the capabilities of Autogluon are assessed. To achieve this, a pipeline is constructed for sktime methods that encompass preprocessing steps and hyperparameter tuning (Section \ref{sktimetuning}). 
The results of tuned sktime methods are then compared with the Autogluon framework.

\section{Background}
%mathematical formulas for time series - meaning univariate, seasonality, drift, one-step-ahead and multi-step-ahead forecasting

\subsection{Time series data}
A time series is a sequence of data points collected at equally spaced intervals in time. It is typically represented as a set of values denoted as $\{ x_t : t = 1, 2, ..., n \}$, which corresponds to data points sampled at discrete time instances 1, 2, ..., n  and the notation $\hat{x}_{t+k|t}$ refers to a forecast made at time t for a future value at time t+k \citep{cowpertwait2009introductory}. A forecast is a prediction of a future value, and the term \say{k} (lead time) indicates how many time steps into the future this prediction is made. Frequently, forecasting involves two distinct approaches: one-step-ahead and multi-step-ahead forecasting. In a one-step-ahead forecast, the primary focus is on predicting just the next value, denoted as  $\hat{x}_{t+1|t}$, where the value of $'k'$ is equal to 1. Conversely, in multi-step-ahead forecasting, the goal extends to predicting a series of future values, $\hat{x}_{t+1|t}$,..., $\hat{x}_{t+h|t}$, with $'h'$ representing the forecast horizon. This forecast-horizon specifies the particular future time points we intend to predict, providing a broader perspective on expected outcomes.

The forecast can either be a point prediction or a prediction interval. A point forecast predicts a specific value or an average value for a future event. For instance, the demand for a given product in the upcoming week will be 10 units. In this case, 10 units is a single, specific value that represents the forecasted demand. In contrast, an interval or probabilistic forecast provides a range of possible values along with associated probabilities. For example, there is a 95 percent chance that the demand for the product in the upcoming week will fall within the range of 10 to 20 units. This forecast provides a broader perspective, indicating that demand is likely to be between 10 and 20 units, with a high level of confidence, rather than specifying a single value.

%The forecast can either be a point prediction or a prediction interval. A point forecast predicts a specific value or an average value for a future event. For instance, the demand for a given product in the upcoming week will be 10 units. In this case, 10 units is a single, specific value that represents the forecasted demand. In contrast, an interval or probabilistic forecast provides a range of possible values along with associated probabilities. For example, there is a 90 percent chance that the demand for the product in the upcoming week will fall within the range of 10 to 20 units. This forecast provides a broader perspective, indicating that demand is likely to be between 10 and 20 units, with a high level of confidence, rather than specifying a single value.
\begin{figure*}[th]
  \centering
  $\begin{array}{cc}
  \includegraphics[width=0.5\linewidth]{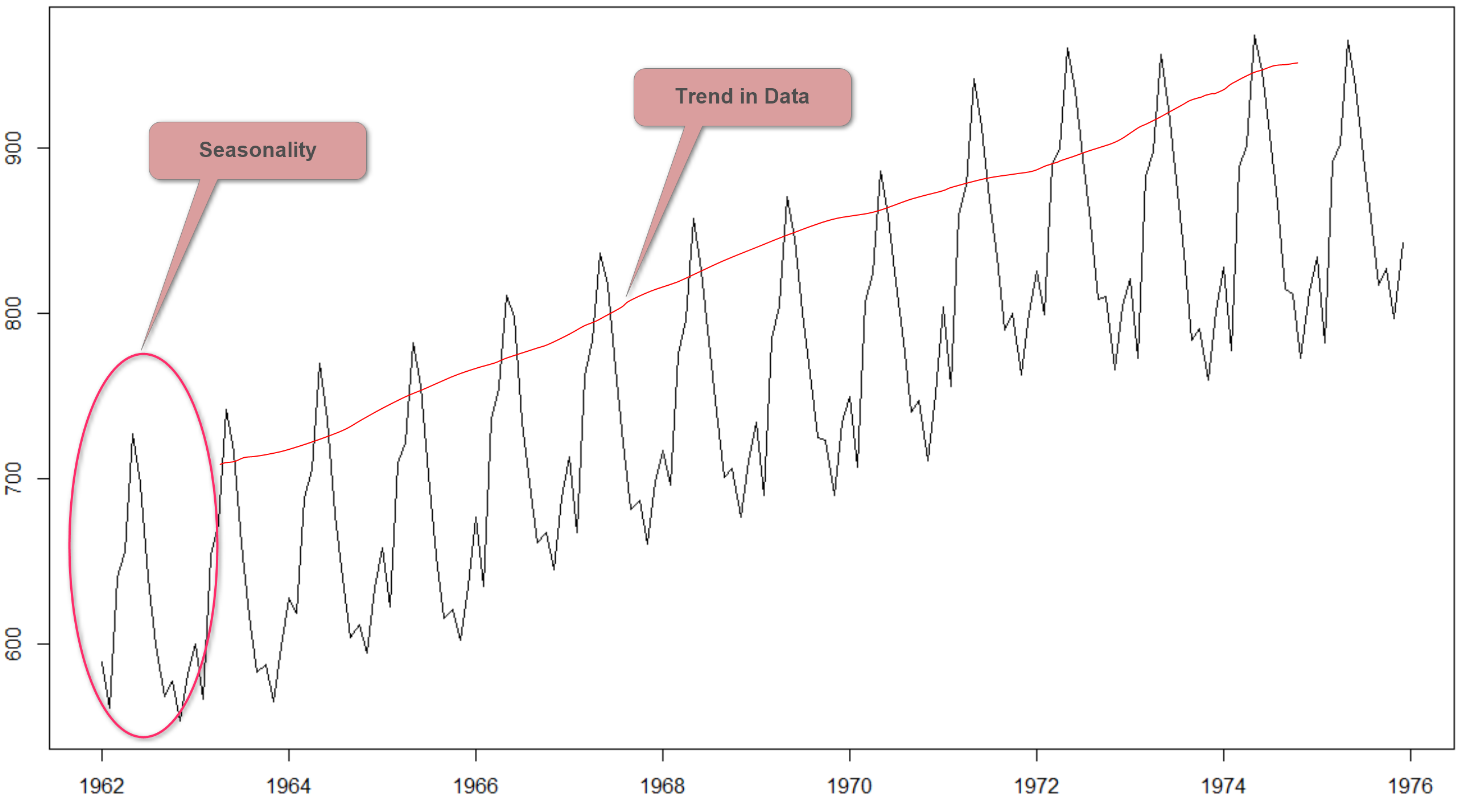} &
  \includegraphics[width=0.5\linewidth, height=1.65in]{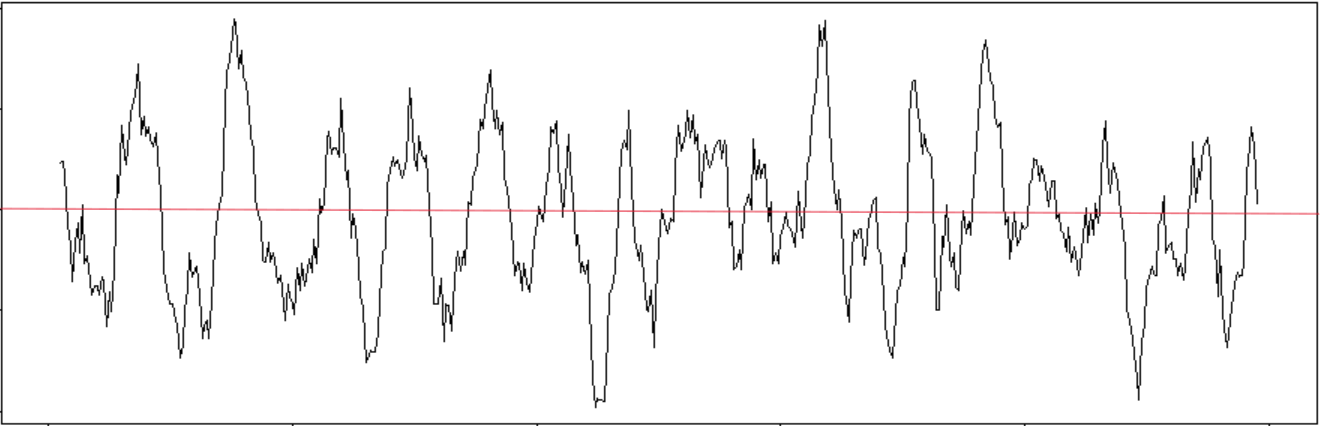} \\
  \mbox{(a)} & \mbox{(b)}  
  \end{array}$
  \caption{\label{Views:Figure1}(a) Trend, Seasonality and (b) Stationarity in a time series data}
\end{figure*}

With the foundational principles of time series data in place, we can now dive deeper into the intricacies of time series analysis. Specifically, we will explore the concepts of trend, seasonality, and stationarity, which play pivotal roles in understanding and modeling time-dependent data patterns. Trend, or recurrent drift, represents a systematic component that evolves over time, showing either linear or nonlinear tendencies without repetitive patterns.  Seasonality, or gradual drift, represents a systematic component that changes over time, typically demonstrating linear or nonlinear behavior but repeating itself at fixed time intervals as shown in \textit{Figure} \ref{Views:Figure1}(a). In summary, a trend pattern emerges when there is a sustained shift in the mean level of the time series, while a seasonal pattern arises when the time series is influenced by seasonal factors like the day of the week or the month of the year \citep{makridakis2008forecasting}. Additionally, the time series data is considered stationary when it lacks either trend or seasonal effects. In practical terms, stationary time series exhibit consistent statistical properties over time, including a stable mean and variance as shown in \textit{Figure} \ref{Views:Figure1}(b). Identifying and incorporating these trends and seasonal patterns in forecasting models is necessary for making informed predictions that align with the underlying patterns in the time series data, ultimately improving the model's predictive accuracy and reliability.

% \begin{figure}[h!]
%     \centering
%     \includegraphics[width=13cm, height=5cm]{trendseasonal.PNG}
%     \caption{Trend and Seasonality in a time series data}
%     \label{fig:ts}
% \end{figure}

% \begin{figure}[h!]
%     \centering
%     \includegraphics[width=13cm, height=5cm]{stationary.PNG}
%     \caption{Stationarity in a time series data}
%     \label{fig:s}
% \end{figure}

\subsection{Frameworks}
%models, methods, concept: explain your models and describe your techniques and methods and why you chose them (depends heavily on the type of project) 
In this section, we will discuss the two time series frameworks that are used in this study.
\subsubsection{AUTOGLUON-TIMESERIES (AG–TS)}
\label{sec:autogluon}

AG–TS \citep{shchur2023autogluon} is a recent open-source AutoML library designed for time series forecasting which offers the flexibility of both local and global forecasting methods. Local forecasting is an approach that involves training an individual predictive model for each time series or fitting classical parametric statistical models to each time series separately. This method is commonly used for capturing patterns like trends and seasonality and employs traditional techniques such as SeasonalNaive, Theta, ARIMA, and ETS. Conversely, global forecasting, leverages multiple time series to train a single \say{global} predictive model or employs advanced machine learning techniques that consider all time series collectively. In the global modeling approach, AG-TS utilizes two main categories of models: deep learning models like DeepAR, PatchTST, and Temporal Fusion Transformer, and tabular models like Recursive Tabular and Direct Tabular. AG-TS also makes use of PyTorch-based deep learning models sourced from GluonTS \citep{alexandrov2020gluonts}.

% AG–TS gives the freedom to choose from four predefined presets that contain these models with default hyperparameter configuration used for fitting. These are namely \say{fast\_training} which contains simple statistical and baseline and fast tree-based models; \say{medium\_quality} which has additional deep learning models like TemporalFusionTransformer; \say{high\_quality} with more powerful deep learning, ml, and statistical forecasting models; and \say{best\_quality} with more cross-validation windows; with an option to manually configure these models. The higher the quality of the presets, the higher the training time but the result is better forecasts. Higher training time can be controlled by specifying the time limit to avoid training all the models until they have been fit. As for hyperparameter tuning AG–TS relies on ensembling techniques instead of HPO or neural architecture search. In ensembling, the individual models are combined using 100 steps of the forward selection algorithm. While AG–TS does support HPO techniques, HPO is excluded from most preset configurations to reduce training time and minimize overfitting on the validation data. There is also a possibility to define custom hyperparameter search spaces for model hyperparameters where a random search is employed to determine the best configuration for the model. 

AG-TS provides four predefined presets to select these models, each containing models with default hyperparameter configurations; with the additional option for manual configuration. These presets include \say{fast\_training}, which incorporates simple statistical, baseline, and fast tree-based models; \say{medium\_quality}, featuring additional deep learning models like TemporalFusionTransformer; \say{high\_quality}, including more powerful deep learning, machine learning, and statistical forecasting models; and \say{best\_quality}, offering more cross-validation windows. The quality of these presets corresponds to the training time and forecast accuracy, with higher-quality presets resulting in better forecasts but longer training times. To manage training time, a time limit can be specified to avoid exhaustive training of all models. AG–TS adopts ensembling techniques for hyperparameter tuning, utilizing 100 steps of the forward selection algorithm %\citep{caruana2004ensemble} 
to combine individual models. Although AG–TS supports Hyperparameter Optimization (HPO) techniques, most preset configurations exclude HPO to reduce training time and minimize overfitting on validation data. Additionally, custom hyperparameter search spaces can be defined for model hyperparameters, employing a random search to determine the optimal configuration for the model.

%In a nutshell, AG-TS performs fit() and predict() methods as follows: The predictor preprocesses the data, fits and evaluates various models using cross-validation, optionally performs hyperparameter optimization (HPO) on selected models, and trains an ensemble of individual forecasting models. After the predictor has been fit, the predict() method is used to generate predictions on new data—including time series that haven’t been seen during training.

\subsubsection*{METHODS}
\label{autogluonmethods}
Below are some of the important methods discussed earlier in detail:

\begin{enumerate}
    \item DeepAR - Autoregressive forecasting model based on a recurrent neural network specifically Long Short-Term Memory (LSTM) networks \citep{salinas2020deepar}. These networks are designed to learn a global model from historical data encompassing all time series within a dataset. The model handles multiple related time series during training by feeding the model with data from different time series, allowing it to learn shared patterns and dependencies across related variables thereby enhancing the forecasting accuracy. DeepAR uses Gaussian likelihood for real-valued data and the negative-binomial likelihood for positive count data as its loss function, along with other likelihood models such as a Bernoulli likelihood for binary data, a beta likelihood for data in the unit interval, or mixtures to handle complex marginal distributions.

    \item PatchTST (Patch time series transformer) - Divides time series data into smaller segments known as \say{patches} and trains on these segments individually before combining their forecasts to generate accurate predictions for the entire dataset as shown below.
    %Nie, Yuqi, et al. “A Time Series is Worth 64 Words: Long-term Forecasting with Transformers.” International Conference on Learning Representations. 2023.    
    \begin{figure}[h!]
    \centering
    \includegraphics[width=0.75\textwidth]{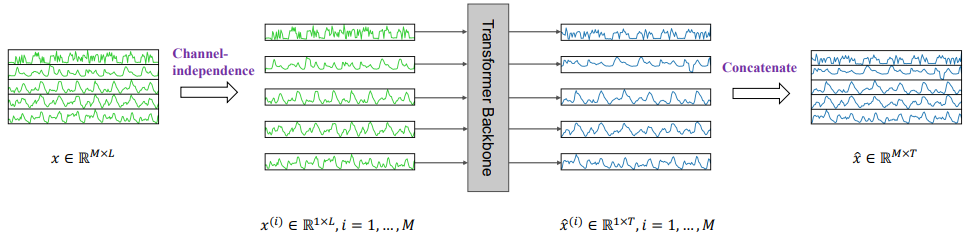}
    \caption{PatchTST model overview from \citep{nie2022time}}
    \label{fig:patchtst}
    \end{figure} 
     %with L:(x1, ..., xL) as the lookback window, where each xt at time step t is a vector of dimension M with T future values (xL+1, ..., xL+T)  

    These patches help the model to extract local semantic meaning and capture meaningful temporal relationships by considering a set of time steps collectively rather than handling the time step individually \citep{marco2023}.
    
    \item Temporal Fusion Transformer (TFT) - An attention-based Deep Neural Network architecture combining LSTM with a transformer layer for multi-horizon time series forecasting \citep{lim2021temporal}. The architecture predicts the quantiles of all future target values as:     
    %Lim, Bryan, et al. “Temporal Fusion Transformers for Interpretable Multi-horizon Time Series Forecasting.” International Journal of Forecasting. 2021.
    \begin{equation*}
        Quantile forecasts = \hat{y}_{t+1}(0.1) \hspace{0.7em} \hat{y}_{t+1}(0.5) \hspace{0.7em} \hat{y}_{t+1}(0.9) \hspace{0.7em} ... \hspace{0.7em} \hat{y}_{t+\tau}(0.1) \hspace{0.7em} \hat{y}_{t+\tau}(0.5) \hspace{0.7em} \hat{y}_{t+\tau}(0.9)
    \end{equation*}
    % given a timestep t, a lookback window k, and a $\tau_{max}$ step ahead window, where t $\in$ [t-k..t+$\tau_{max}$], with observed past inputs x $\in$ [t-k..t], future known inputs x $\in$ [t+1..t+$\tau_{max}$], a set of static variables \say{s} and the target variable \say{y} $\in$ [t+1..t+$\tau_{max}$]. During training, TFT is optimized by jointly minimizing the quantile loss, which is the sum of losses across all these quantile outputs.  
    where t $\in$ [t-k..t+$\tau_{max}$] is a timestep, k is a lookback window, $\tau_{max}$ is a step-ahead window, and the target variable y $\in$ [t+1..t+$\tau_{max}$]. The model utilizes observed past inputs x $\in$ [t-k..t], future known inputs x $\in$ [t+1..t+$\tau_{max}$], a set of static variables \say{s}, variable selection networks (selects relevant features at each time step), LSTMs (handles local processing), Gated Residual Networks (enables efficient information flow), and multi-head attention (integrates information from any time step) for prediction. Throughout the training process, TFT optimizes its performance by jointly minimizing the quantile loss defined as follows:
    \begin{equation}
        QL(y,\hat{y},q) = max[q(y-\hat{y}),(1-q)(y-\hat{y})]
    \end{equation}

    where $q$ $\in$ [0,1], $y$ is the actual value, and $\hat{y}$ is the predicted value. 
    
    \item Recursive and Direct Tabular models - 
    A recursive tabular model predicts future time series values sequentially, one at a time, building upon the outcomes of the previous steps through a recursive process, as shown below:
    \begin{figure}[h!]
    \centering
    \includegraphics[width=0.75\textwidth]{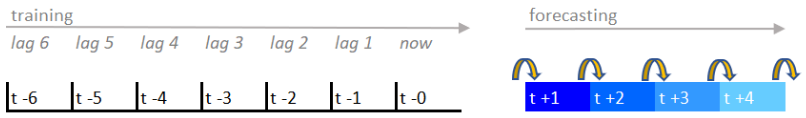}
    % \caption{Recursive Tabular model  from \href{https://github.com/cerlymarco/tspiral#overview}{\textit{Github}}
    \label{fig:patchtst}
    \end{figure}  
    
    On the other hand, a direct tabular model predicts all future time series values simultaneously, taking into consideration each time step to be forecast as shown below:
    \begin{figure}[h!]
    \centering
    \includegraphics[width=0.75\textwidth]{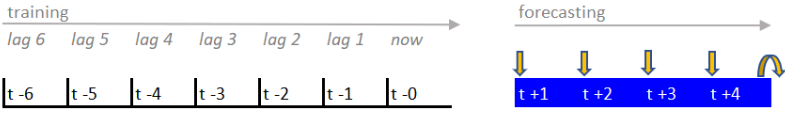}
    % \caption{Direct Tabular model from \href{https://github.com/cerlymarco/tspiral#overview}{\textit{Github}}
    \label{fig:patchtst}
    \end{figure}

%https://joaquinamatrodrigo.github.io/skforecast/0.4.3/notebooks/autoregresive-forecaster.html
%https://towardsdatascience.com/how-to-improve-recursive-time-series-forecasting-ff5b90a98eeb

    %RecursiveTabular predicts future time series values one by one. DirectTabular, on the other hand, predicts all future time series values simultaneously.
    %or employs recursive forecasting techniques that iteratively update the model by gradually incorporating new data, leading to refined forecasts. 
    
\end{enumerate}

% One of the notable strengths of GluonTS lies in its capacity to facilitate probabilistic modeling for large collections of time series. It accomplishes this by predicting the probability distribution (p) of future values, taking into account past values, covariates, and model parameters.

%https://towardsdatascience.com/local-vs-global-forecasting-what-you-need-to-know-1cc29e66cae0
%add this 

\subsubsection{SKTIME}
\label{sec:sktime}

Sktime \citep{loning2019sktime} is an open-source Python library designed for time series data that supports various time series tasks, including forecasting, classification, regression, clustering, and annotation. While other machine learning libraries can be used for time series, sktime distinguishes itself as a dedicated tool tailored specifically for time series analysis. Unlike general-purpose machine learning libraries like scikit-learn (sklearn), sktime is purpose-built for working with time series data. It offers specialized functionalities that streamline the complexities associated with handling temporal data, mitigating risks related to manual coding, addressing data leakage concerns, and tackling multi-step forecasting challenges making it a better fit for time series forecasting and analysis.

\subsubsection*{METHODS}
\label{sktimemethods}
The library includes a variety of algorithms for classification, regression, clustering, and forecasting. Below are some of the \href{https://www.sktime.net/en/stable/api_reference/forecasting.html}{\textit{forecasting}} methods that are considered for this study.

% the forecasting methods that are considered for this research project are as follows:

\begin{enumerate}
    \item Naive - The simplest forecasting technique, relies on the assumption that past trends will persist into the future. It entails using the actual observed values from the most recent time period as the forecast for the next period as follows: 
    \begin{equation}
        x_{t+i|t} =  x_{t-k+i}
    \end{equation}
  
    %https://otexts.com/fpp2/simple-methods.html
    
    This method is pretty straightforward and doesn't incorporate complex predictive models, external factors, or any form of adjustment based on additional information. %[\href{https://www.sktime.net/en/stable/api_reference/auto_generated/sktime.forecasting.naive.NaiveForecaster.html}{\textit{sktime-naive}}]. %google and sktime

    \item STL (Season-Trend decomposition using LOESS (Locally Weighted Scatterplot Smoothing)) - Decomposes a time series into three components: trend, season(al), and residual where the trend component characterizes the long-term, underlying behavior within the time series; the seasonal component captures the recurrent, periodic fluctuations in the data; and the residual component, also known as the remainder or error, represents the unexplained and irregular fluctuations in the data. 
    The forecast, \say{$y_{\text{pred}}$}, is obtained by summing these three components as follows: %.[\href{https://www.sktime.net/en/stable/api_reference/auto_generated/sktime.forecasting.trend.STLForecaster.html}{\textit{sktime-stl}}].
    \begin{equation}
        y_{\text{pred}} = y_{\text{pred\_trend}} + y_{\text{pred\_seasonal}} + y_{\text{pred\_residual}}
    \end{equation}
    
    \item Theta -  %[\href{https://www.sktime.net/en/stable/api_reference/auto_generated/sktime.forecasting.theta.ThetaForecaster.html}{\textit{sktime-theta}}] 
     Revolves around adjusting local trends within a time series by introducing a critical factor known as the Theta-coefficient ($\theta$). This is then directly applied to the changes in values over time (second differences) of the time series as follows:    
     \begin{equation}
         X''_{\text{new}}(\theta) = \theta \cdot X''_{\text{data}}
     \end{equation}
     where $X''_{\text{data}}$ = $X_{t}$ - 2$X_{t-1}$ + $X_{t-2}$ at time t\citep{assimakopoulos2000theta}. 
     %The smaller the value of $\theta$, the larger the degree of declination as shown in Figure \ref{fig:theta}. 
     The model breaks down the original time series into two or more separate Theta lines which then undergo a fitting process, where predictions are made using a Simple Exponential Smoother. The forecasts produced for each line are then combined to produce a final forecast.
    % \begin{figure}[h!]
    % \centering
    % \includegraphics[width=0.75\textwidth, height=0.15\textheight]{theta.PNG}
    % \caption{Theta model}
    % \label{fig:theta}
    % \end{figure}

    % The model is based on the concept of modifying the local curvatures of the time series. This change is obtained from a coefficient, called Theta-coefficient (as a symbol is used the Greek letter Theta), which is applied directly to the second differences of the time series. In other words, the Theta model method involves decomposing the initial time series into two or more Theta lines and fitting these lines, forecasting the lines using a Simple Exponential Smoother, and then combining the forecasts from the two lines to produce the final forecast.

    \item Trend - Generates predictions for time series data by leveraging trends that are either short-term (hourly, daily, weekly, monthly) or long-term (yearly, decade) patterns within the dataset. The forecasting involves regressing time series values against their corresponding indices, to predict future values as shown below. %\textit{Figure} \ref{fig:Trend}.
    \begin{figure}[h!]
    \centering
    \includegraphics[width=0.4\textwidth, height=0.175\textheight]{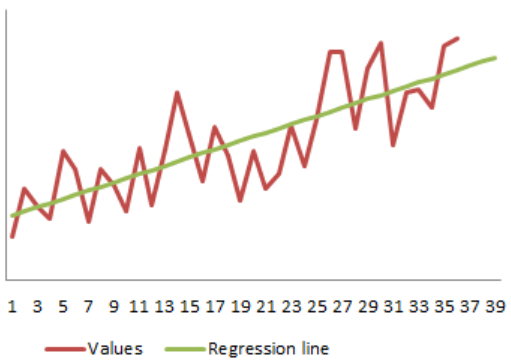}
    \caption{Trend forecasting using Linear regression}
    \label{fig:Trend}
    \end{figure}    
    
    Mathematically, for a given time series ($v_i$, $t_i$), where $v_i$ and $t_i$ denote the values and their corresponding timestamps with $i$ ranging from 1 to T, future predictions are made by fitting a regression model $f$ with an error term $e_i$ as follows:
    \begin{equation}
        v_{i} = f(t_{i}) + e_{i} 
    \end{equation}
    %  This model captures the underlying trend, and for any new time point $t_*$, the forecasted value $f(t_*)$ can be predicted.
    % \todo{check this again later}
    
    %Generates forecasts for time series data based on trends. It can be either for short-term (month) or long-term (decade) trends in the data. One of the common ways for trend-based forecasting is through regression on time indices of the time series data. For a given time series (vi, ti), i=1,...T, where vi are the values and ti are time stamps, a regression model vi=f(ti)+ei is fitted, and for any new time point t\_*, f(t\_*) is predicted.

    %by performing regression on time series values against their corresponding index.  It employs a sklearn regressor, which is determined by the \say{regressor} parameter, to conduct the regression analysis on time series values relative to their respective indices, ultimately producing forecasts that are trend-driven. 
    %[\href{https://www.sktime.net/en/stable/api_reference/auto_generated/sktime.forecasting.trend.TrendForecaster.html}{\textit{sktime-trend}}]. %from sktime
    
    \item Polynomial Trend - Generates predictions for time series data by leveraging trends within the dataset similar to the Trend method with an additional step of polynomial feature extraction. Mathematically, for a given time series ($v_i$, $p(t_i)$), where $v_i$ and $t_i$ denote the values and their corresponding timestamps with $i$ ranging from 1 to T, and $p$ representing the polynomial feature transform with a degree and an intercept, future predictions are made by fitting a regression model $f$ with an error term $e_i$ as follows:
    \begin{equation}
        v_i = f(p(t_i)) + e_i 
    \end{equation}

    % Generates forecasts for time series data based on polynomial trends. This is similar to the Trend method, with an additional step of polynomial feature extraction. For a given time series ($v_i$, $p(t_i)$), where $v_i$ and $t_i$ denote the values and their corresponding timestamps (for i ranging from 1 to T), and p represents the polynomial feature transform with a degree and an intercept, a regression model can be formulated as:
    % \begin{equation}
    %     v_i = f(p(t_i)) + e_i 
    % \end{equation}
    %  This model captures the underlying polynomial trend, and for any new time point $t_*$, the forecasted value $f(p(t_*))$ can be predicted.
    % \todo{check this again later}

    %This method also leverages a sklearn regressor determined by the \say{regressor} parameter to carry out regression on time series values in relation to their associated indices, following the application of polynomial feature extraction. It's similar to the Trend Forecaster, but in this case, the regressor is integrated with a transformation step that involves Polynomial Features applied to the time index at the outset. %[\href{https://www.sktime.net/en/stable/api_reference/auto_generated/sktime.forecasting.trend.PolynomialTrendForecaster.html}{\textit{sktime-polytrend}}]. %from sktime    
    
    \item AutoRegressive Integrated Moving Average (ARIMA): 
    %This is a StatsForecast AutoARIMA estimator that identifies the most suitable ARIMA model based on either the BIC, AICc, or AIC  criteria. %[\href{https://www.sktime.net/en/stable/api_reference/auto_generated/sktime.forecasting.statsforecast.StatsForecastAutoARIMA.html}{\textit{sktime-arima}}]. where predictions are based on the weighted linear sum of recent past observations or lag 
    The modeling approach with three main components: AR (AutoRegressive), I (Integrated), and MA (Moving Average); \say{AR} capturing the relationship between an observation and a certain number of past observations; \say{I} differencing raw observations to make the time series stationary; and \say{MA} handling the dependency between an observation and the residual error from a moving average model applied to lagged observations. In the standard notation, ARIMA(p, d, q), where p, d, and q represent the order, degree, and order of the AutoRegressive, Integrated, and Moving Average components, respectively, a simple ARIMA(1,1,1) model can be defined as:
    %The standard notation for the ARIMA model is ARIMA(p,d,q) where p, d, and q represent the parameters. For a simple model ARIMA(1,1,1), the general form of the model can be written as:
    \begin{equation}
        \Delta Y_t = c + \phi_1 \Delta Y_{t-1} +\theta_1\varepsilon_{t-1} +\varepsilon_t
    \end{equation}
    where,
    \begin{itemize}
        \item c is a baseline constant factor
        \item $Y_t$ and $Y_{t-1}$ are the  values of the time series in the current period (t) and the previous period $(t-1)$, respectively      
        \item $\varepsilon_t$ and $\varepsilon_{t-1}$ are the error terms for the current period (t) and the previous period $(t-1)$, respectively
        \item $\Delta Y_{t-1}$ = $Y_{t-1}$ - $Y_{t}$ is the difference between the values in the period $t-1$ and t; $\Delta Y$ is the entire time series representing the disparity between consecutive periods
        \item $\phi_1$ is the autoregressive parameter of lag 1 that expresses the impact of the lagged difference between values ($\Delta Y_{t-1}$) on the current difference ($\Delta Y_{t}$) 
        \item $\theta_1$ is the moving average parameter of lag 1 that represents the impact of the lagged error term ($\varepsilon_{t-1}$) on the current difference ($\Delta Y_{t}$) \citep{viktor2023}.
    \end{itemize}

\textit{Note:} In our study, we employ the StatsForecastAutoARIMA estimator, which automatically selects the best ARIMA model using either AIC, AICc, or BIC values.
    
    %https://365datascience.com/tutorials/python-tutorials/arima/
    
    %denotes the lag order, representing the number of lag observations incorporated in the model; d represents the degree of differencing, denoting the number of times raw observations undergo differencing; and q denotes the order of moving average, indicating the size of the moving average window.
    
    %The StatsForecastAutoARIMA function performs a search among potential models within the specified order constraints.%sktime and %https://machinelearningmastery.com/arima-for-time-series-forecasting-with-python/
    
    \item ExponentialSmoothing - %[\href{https://www.sktime.net/en/stable/api_reference/auto_generated/sktime.forecasting.exp_smoothing.ExponentialSmoothing.html}{\textit{sktime-expo}}] 
    Generates predictions for a time series by computing weighted averages of previous observations where the weights diminish exponentially as the observations become more dated. It is defined as:
    \begin{equation}
        s_{(t)} = \alpha x_{(t)} + (1-\alpha)s_{(t-1)}
    \end{equation}
    where, \begin{itemize}
        \item $s_{(t)}$ is the smoothed value at time t
        \item $s_{(t-1)}$ is the former smoothed statistic
        \item $x_{(t)}$ is the observed value at time t 
        \item $\alpha$ is the smoothing factor of data between 0 (lower weights to recent changes) and 1 (higher weights to recent changes) \citep{expo2023}.
    \end{itemize}
    There are three primary variations of this method: Single Exponential Smoothing (SES), which is used when there is no inherent trend or seasonality in the time series; Double Exponential Smoothing, which explicitly accommodates trends ($\beta$) in the time series; and Triple Exponential Smoothing which incorporates seasonality ($\gamma$) within the time series.

    %https://www.influxdata.com/blog/exponential-smoothing-beginners-guide/
    
    %methods are similar in that a prediction is a weighted sum of past observations, but the model explicitly uses an exponentially decreasing weight for past observations. Specifically, past observations are weighted with a geometrically decreasing ratio. Forecasts produced using exponential smoothing methods are weighted averages of past observations, with the weights decaying exponentially as the observations get older. In other words, the more recent the observation the higher the associated weight.
    %https://machinelearningmastery.com/exponential-smoothing-for-time-series-forecasting-in-python/
    %https://machinelearningmastery.com/how-to-grid-search-triple-exponential-smoothing-for-time-series-forecasting-in-python/ - for tuning
 
    \item ETS (Error, Trend, Seasonality) - Generates forecasts by considering combinations of additive or multiplicative error models, additive or multiplicative trend components, and additive or multiplicative seasonality factors. These models are integral components of exponential smoothing methods and the most basic ETS model, commonly known as simple exponential smoothing, is denoted by the (A, N, N) model notation. This model is characterized by additive errors, no trend, and no seasonality.
    \\ \textit{Note:} In our study, we employ the StatsForecastAutoETS estimator, which automatically selects the best ETS model using Maximum likelihood estimation (MLE) and AICc.
    
    %https://otexts.com/fpp2/ets.html
    
    %chosen information criterion (like AICc) and estimation method like Maximum likelihood estimation (MLE)
    
    %This is also a StatsForecast AutoETS estimator which is used for the automatic selection of the most appropriate ETS (Error, Trend, Seasonality) model based on a chosen information criterion (like AICc) and estimation method like Maximum likelihood estimation (MLE). %[\href{https://www.sktime.net/en/stable/api_reference/auto_generated/sktime.forecasting.statsforecast.StatsForecastAutoETS.html}{\textit{sktime-ets}}].
    %sktime %https://nixtla.github.io/statsforecast/docs/models/autoets.html
    
    %https://medium.com/analytics-vidhya/time-series-forecasting-models-726f7968a2c1
 
    \item Prophet -  %[\href{https://www.sktime.net/en/stable/api_reference/auto_generated/sktime.forecasting.fbprophet.Prophet.html}{\textit{sktime-prophet}}] 
    Predicts time series data by employing an additive model that captures non-linear trends, seasonality patterns, as well as the influence of holidays similar to Holt-Winters or triple exponential smoothing. The additive regression model typically uses a piecewise linear or logistic growth curve trend and the Fourier series (yearly) or dummy variables (weekly) for seasonal component modeling \citep{sean2017}. By default, the model is engineered to automatically handle the majority of the forecasting process and works best when applied to time series with prominent seasonal variations and an extensive history of data.
    %(\href{https://facebook.github.io/prophet/}{\textit{Prophet}}). 
 
    % \item RecursiveTabularRegression and DirectTabularRegression - Both of them are used to simplify the transition from forecasting to tabular regression. %[\href{https://www.sktime.net/en/stable/api_reference/auto_generated/sktime.forecasting.compose.RecursiveTabularRegressionForecaster.html}{\textit{sktime-Tabular}}]. 
    % These forecasters can be implemented using a variety of models, such as LightGBM, LinearRegression, etc. In our use case, we employ them to incorporate the LightGBM model.
    % %, along with global pooling, to establish a global model.
   
\end{enumerate}

\section{Related Work}
%Review of existing literature and studies on AutoML for time series data. Discussion of the strengths and limitations of previous approaches.Identification of gaps or areas for improvement in the field.
% Comparing different methods to assess their performance has been done for many years now in the field of machine learning. From comparing two methods on one dataset to comparing n methods with n datasets, researchers have come a long way. Researchers have made continuous efforts to evaluate various models using diverse time series data to identify and enhance both the theory and application of different time series tasks. This ongoing effort is dedicated to uncovering insights and advancing both the theoretical understanding and practical applications of different time series tasks.
In this section, we delve into existing research and work in the field of benchmarks for time series analysis. Over time, the practice of comparing the effectiveness of various methods has become integral to machine learning, aiding in the selection of the most suitable model for any given task. This evolution has moved from comparing two methods on a single dataset to evaluating n methods across diverse datasets spanning different domains.

Within the domain of time series analysis, researchers have consistently sought to assess different time series models using a variety of time series data to find the best model. To conduct a thorough evaluation of these models, a benchmark is typically employed. Benchmark can be defined as a standardized set of tasks, datasets, or problems that are used to assess and compare the performance of different algorithms, models, or systems \citep{leenings2022recommendations}.  In other words, the benchmark refers to a tool for systematically assessing and comparing methods or systems based on specific attributes, such as performance \citep{v2015build}. It serves as a yardstick for evaluating the capabilities of various approaches consistently and objectively.

\subsection{Time series forecasting}
One of the notable studies for benchmarking time series forecasting methods is the M4 Competition \citep{makridakis2020m4}. This competition was built upon the insights gained from three prior M competitions, all with the overarching goal of advancing both the theoretical understanding and practical applications of different time series forecasting techniques. All the M competitions aimed to improve forecasting accuracy through hands-on experience and real-world data analysis. The M4 competition, in particular, employed 12 benchmark methods namely statistical benchmarks including theta method, and different variations of naive and exponential smoothing methods; ML benchmarks including MLP, RNN, and combinations of SES, Holt, and Damped exponential smoothing methods; and standard methods like ETS and ARIMA for comparative analysis. The evaluation was conducted on 100,000 time series data points from the ForeDeCk database, involving 49 submissions from leading forecasters. Although the study provided a solid foundation for drawing concrete conclusions and deriving valuable insights from empirical evidence, the analysis was rather limited to fewer datasets and generic forecasting algorithms. 

Similarly, in \say{Libra: A Benchmark for Time Series Forecasting Methods} by \citep{bauer2021libra}, various forecasting methods were compared to identify the most suitable method for specific use cases. The authors conducted the analysis using 400 real-world time series, employing classical forecasting models and regression-based machine learning models, amounting to a total of 10 methods for evaluation. These datasets were categorized into four distinct use cases: Economics, Nature and Demographics, Finance, and Human Access to evaluate and rank time series forecasting techniques based on their performance across a diverse set of evaluation scenarios. While the results and analysis were promising, the methods were assessed under default settings without parameter tuning, limiting the scope of the findings.

\subsection{Time series classification and clustering}
In a similar context, ongoing research has delved into benchmarking time series tasks focusing on classification and clustering as well. For instance, the authors in \citep{pfisterer2019benchmarking} systematically evaluated various classification algorithms for functional data and machine learning approaches to identify the most efficient solutions for time series classification problems. The evaluation was conducted on approximately 51 datasets of small and medium sizes from the UCR archive, featuring a diverse range of classes, and employed 80 algorithms both with default and tuned settings. Interestingly, hyperparameter tuning of the machine learning algorithms did not lead to a significant improvement in performance. Thus, leaving room for the incorporation of new methodologies and ensemble techniques to enhance performance.

Likewise, in \citep{JAVED2020100001}, a benchmark study was conducted to evaluate and rank the performance of different time series clustering algorithms using diverse datasets. The evaluation involved three categories of clustering algorithms namely partitional, density-based, and hierarchical totaling 8 general-purpose clustering methods and three distance measures namely Euclidean, dynamic time warping (DTW), and shape-based, across 112 datasets from the UCR archive. The results were ranked based on the number of winning performances across all datasets providing a comprehensive comparative analysis. Nonetheless, giving an opportunity to delve further into exploring additional clustering algorithms and conducting evaluations based on domain-specific datasets.

%The results showed the overall performance of all eight algorithms to be similar with high sensitivity to the datasets, indicating that no method is superior to the others for all datasets.
%The findings revealed a comparable overall performance among all eight algorithms, with high sensitivity to the datasets, suggesting that no single method exhibited superiority across all datasets.

%The authors ranked the performance of each clustering method and recorded the number of winning performances across all available datasets. The objective was to offer a comparative analysis with established methods, providing accessible and reusable results for other researchers.

%The datasets used in the study are limited to general-purpose clustering methods rather than domain-specific.

While these studies have made substantial contributions to the benchmarking field in time series analysis, particularly in time series forecasting, conducting a comparative analysis between AutoML tools such as AutoGluon-Timeseries and the methodologies within a robust framework like sktime represents a novel approach. This comparison allows us to explore how automation in time series forecasting, may yield improvements or drawbacks, providing valuable insights.

\section{Methodology}

%\subsection{Benchmarking}
\label{sec:benchmarking}
%Your method section should be about *your* approach to benchmarking. Again, you can look at the AutoML benchmark paper attached for inspiration on the structure.

In this section, we will discuss the benchmark datasets, evaluation metrics, and the experimental setup used in this study. 

\subsection{Benchmark datasets}
\label{sec:benchmarking}
%In general, a benchmark is a standardized set of tasks, datasets, or problems that are used to assess and compare the performance of different algorithms, models, or systems. It serves as a yardstick for evaluating the capabilities of various approaches consistently and objectively. In other words, the benchmark refers to an instrument used to evaluate and/or compare systems or methods based on certain properties \citep{bauer2021libra}. 

To evaluate different frameworks based on their performance, the selection of a robust set of benchmark datasets is important. As outlined in Section \ref{sec:introduction} and different from previous benchmarks, the datasets used in this study have been sourced from the Monash Time Series Forecasting Repository, which not only showcases a wide range of diversity in datasets but also encompasses different domains such as tourism, banking, energy, economics,  transportation, nature, web,  sales, and health. The number of time series per dataset varies from as few as one to as many as 145,063 and the length of these time series ranges from a minimum of 11 data points to a maximum of 7,397,147 data points. These datasets are in the .tsf format and include time series of both fixed and variable lengths. All these datasets are readily accessible on \href{https://forecastingdata.org/}{\textit{Zenodo}}. The details of these datasets along with their frequency and forecast horizon (fh) are given in Appendix \ref{app:theorem}.

%M4-100000  Wind Power-7397147 – change this in the report
%Web traffic - 145,063 solar power - 7,397,222

\subsection{Evaluation metrics}
\label{sec:evaluationmetrics}

To compare different methods, the choice of evaluation metrics is of paramount importance. However, finding error measures that consistently perform well across all forecasting models is a daunting challenge and is heavily dependent on the specific domain and use cases. Among the numerous available metrics, sMAPE, RMSE, and MASE are widely recognized and commonly used point forecast metrics in the field of time series forecasting. A point forecast is selected as opposed to probabilistic forecasting due to the limited availability of suitable probabilistic metrics. All three metrics address certain aspects or challenges in forecasting, such as sMAPE handles cases where actual and predicted values can be zero or near zero, RMSE emphasizes the magnitude of errors with sensitivity to outliers, and MASE is resilient to outliers and suitable for comparing models across different scales. However, as a result of varying calculations within AutoGluon and sktime frameworks for RMSE with possible differences across multiple time series with different magnitudes, the study refrains from directly comparing the obtained results. Instead, these findings are provided in tables \ref{table:8} and \ref{table:9} of the appendix. 

%Point forecast metrics are considered as opposed to probabilistic forecast metrics due to its limited availability. 
%These metrics being scaled and scale-free, makes the results comparable and applicable in diverse contexts. 
%These are represented as non-negative floating points, with the best possible value being 0.0 (lower the better).

%\todo{for results check below- commented}
%Absolute and relative metrics measure different aspects of the prediction. So one model is not better than the other in absolute sense (pun intended).Which metric to value depends on your application. When the outcome range is wide (probably your case) and skewed, relative error measurements are better than absolute error measurements.
%https://datascience.stackexchange.com/questions/37168/high-rmse-and-mae-and-low-mape

%Amidst the plethora of available metrics, we have distilled our focus to the following set of point forecast metrics which are scaled and scaled-free metrics and are non-negative floating points with the best possible value being 0.0.  

%https://medium.com/analytics-vidhya/time-series-forecasting-a-complete-guide-d963142da33f

%These metrics are employed to assess the accuracy and efficiency of forecasting methods. Considering all the available metrics out there, we have narrowed it down to following metrics for point forecasts in order to gauge the precision and effectiveness of forecasting methods.

\begin{enumerate}
    \item \textbf{Symmetric mean absolute percentage error (sMAPE/SMAPE): }    %https://scholarworks.utep.edu/cgi/viewcontent.cgi?article=1865&context=cs_techrep
     Is an error metric based on relative errors and is an improvement over the traditional Mean Absolute Percentage Error (MAPE) accuracy measure. It is a balanced metric, which considers both overestimation and underestimation errors and handles cases where both the forecasted and actual values can be zero or near zero. Lower sMAPE values indicate better model performance, with the best possible value being 0.0. It is defined as follows:

    % \begin{equation}
    %     \frac{1}{n} \sum_{i=1}^{n} |\frac{(Y_{actual} -Y_{predicted})}{Y_{actual}}|
    % \end{equation}

    \begin{equation}
        \frac{2}{n} \sum_{i=1}^{n} \frac{|Y_{actual} -Y_{predicted}|}{|Y_{actual}| + |Y_{predicted}|}
    \end{equation}

    % \item \textbf{MSE and RMSE}

    % Mean Squared Error (MSE) is a representation of how forecasted values differ from actual or true ones. It is represented by the following formula:

    % \begin{equation}
    %     \sum_{i=1}^{n} \frac{1}{n} {(Y_{actual} -Y_{predicted})}^2
    % \end{equation}

    % \item \textbf{RMSE}
    
    % Root Mean Squared Error (RMSE) is the square root of Mean Squared Error \citep{puja2021} and is an error measure that is helpful in situations where it's essential to understand the typical size of errors in predictions as it not only measures the magnitude of errors but also penalizes larger errors more heavily than smaller ones. It is sensitive to outliers, meaning it can be strongly influenced by a few extreme errors in the data. It is as follows:

    % \begin{equation}
    %     \sqrt{\sum_{i=1}^{n} \frac{1}{n} {(Y_{actual} -Y_{predicted})}^2}
    % \end{equation} 
    
    \item \textbf{Mean Absolute Scaled Error (MASE): }
    %https://www.statisticshowto.com/mean-absolute-scaled-error/
     Is a scale-independent error metric that expresses each error as a ratio relative to the average error of a baseline model (Naive). It is resilient to outliers, making it less susceptible to the impact of extreme values, and is well-suited for time series data with seasonality and trends, as it considers the error relative to the persistence forecast. Lower MASE values indicate better model performance, with the best possible value being 0.0. It is defined as follows:
    %In practice, a MASE value less than 1 indicates that the forecasting model is performing better than the naive forecast, while a MASE value greater than 1 suggests that the model is less accurate than the naive forecast. 
    \begin{equation}
        mean ( | \frac{e_t}{\frac{1}{n-m} \sum_{i=m+1}^{n} |Y_{actual} - Y_{actual - m}| } | )
    \end{equation}        
\end{enumerate}
where, 

%$Y_{actual}$ represents the actual or true value, $Y_{predicted}$ denotes the predicted value, n is the number of observations, $e_t$ is the forecast error for a given period, and m is the seasonal period. 

\begin{itemize}
    \item $Y_{actual}$ represents the actual or true value
    \item $Y_{predicted}$ denotes the predicted value  % \item t is the set of forecasting periods ranging from 1 to n
    \item $e_t$ denotes the forecast error for a given period    
    \item $n$ and $m$ represent the number of observations and seasonal period respectively.
\end{itemize}

\subsection{Experiment Setup}
\subsubsection{Framework Configuration}
For AutoGluon, presets are set to \say{best\_quality} as opposed to other presets discussed in Section \ref{sec:autogluon} with the time limit set to 600 seconds and 3600 seconds, to prioritize both high-quality forecasts and robustness, with all other configurations kept at default. This preset contains simple statistical models like ETS, Theta, Naive, and SeasonalNaive, fast tree-based models like RecursiveTabular, and DirectTabular, deep learning models like DeepAR with multiple copies, TemporalFusionTransformer, and PatchTST, automatically tuned statistical models like AutoETS, AutoARIMA, along with additional tabular models. As for sktime, Naive, STL, Theta, Trend, PolynomialTrend, StatsForecastAutoARIMA, ExponentialSmoothing, StatsForecastAutoETS, and Prophet as discussed in Section \ref{sktimemethods} are used with default hyperparameter configurations. For methods supporting the \say{seasonal period} (sp) argument, it is set to the frequency of the data. To ensure a fair comparison, a standardized time limit of 3600 seconds is imposed on all sktime forecasters. 

Both frameworks use multi-step ahead forecasting to predict a series of future values and the train-test split strategy is set on this forecast horizon. Each framework employs its designated method for this purpose, with AutoGluon using the \say{train\_test\_split} method and sktime using the \say{temporal\_train\_test\_split} method. The evaluation metrics used to assess the performance of these frameworks are SMAPE, and MASE as discussed in Section \ref{sec:evaluationmetrics}. Additionally, when optimizing the MASE metric, the \say{eval\_metric\_seasonal\_period} argument in AutoGluon and the \say{sp} argument in sktime are set to the data's frequency to ensure correct MASE computation.

%The test score is computed using the last prediction_length=48 timesteps of each time series in test_data - autogluon

%These serve as a baseline for the purpose of performance comparison. - sktime

%Recognizing that achieving high accuracy isn't the sole crucial factor, it's often essential in practical scenarios to ensure that predictions can be generated within a reasonable timeframe. Hence, 

\subsubsection{Tuning Configuration for sktime Methods}
\label{sktimetuning}
\begin{table}
\centering
\begin{tabular}{|p{4cm}|p{3.6cm}|p{7cm}|}
\hline
\textbf{Forecaster} & \textbf{Parameter} & \textbf{Options} \\
\hline
\multirow{3}{*}{Naive} & sp & \\
& strategy & [last, mean, drift] \\
\hline
\multirow{6}{*}{STL} & sp & \\
& seasonal\_deg & [0, 1, 2] \\
&trend\_deg & [0, 1, 2] \\
& seasonal\_jump & [1, 2, 3] \\
&trend\_jump & [1, 2, 3] \\
& robust & [True, False] \\
\hline
\multirow{2}{*}{Theta} & sp & \\
& deseasonalize & [True, False] \\
\hline
\multirow{1}{*}{Trend} & regressor & [LinearRegression(), Ridge(), SGDRegressor(), RandomForestRegressor()] \\
\hline
\multirow{2}{*}{PolynomialTrend} & regressor & [LinearRegression(), Ridge(), SGDRegressor(), RandomForestRegressor()] \\
& degree & [1, 2, 3] \\
\hline
\multirow{5}{*}{ExponentialSmoothing} & sp & \\ 
& smoothing\_level & [0.1, 0.2, 0.3] \\
& smoothing\_trend & [0.1, 0.2, 0.3] \\
& damping\_trend & [0.2, 0.3, 0.4] \\
& initialization\_method & [heuristic, legacy-heuristic, estimated] \\
\hline
\end{tabular}
\caption{Forecaster parameters used for model tuning}
\label{table:tuning}
\end{table}

To evaluate the comparative performance of the tuned methods within the sktime library against the AutoGluon framework, we have implemented the following tuning process. The process involves a pipeline that incorporates preprocessing steps, including Standard scaling, and box-cox transformations; and hyperparameter tuning of the model's settings as given in Table \ref{table:tuning}, to enhance its performance. Sktime framework offers two tuning meta-forecastors namely ForecastingRandomizedSearchCV and ForecastingGridSearchCV for this purpose. We have employed a random search approach, which involves exploring a range of hyperparameter combinations randomly, rather than exhaustively testing every possible combination as in grid search. This random search method is preferred due to its efficiency and suitability for larger datasets, as it can yield accurate results and, in some cases, even outperform the more computationally intensive grid search.

%Hyperparameter tuning is defined as a process of optimizing the settings of a machine-learning model for any dataset, with an aim to enhance its performance. 

\subsubsection{Hardware}
All experiments are run on HPC Hardware with the following configuration: CPU: 2x Intel Xeon 4110 @ 2.1Ghz (32 hyperthreads) ; RAM: 384GB;
GPU: 4x ASUS Turbo GeForce GTX 1080 Ti (11GB RAM, 3584 CUDA cores, compute capability 6.1). All the jobs are run for a maximum of 24 hours.

% High accuracy is not the only important property of an AutoML system—the ability to generate predictions in a reasonable amount of time is often necessary in practice. To evaluate the efficiency of AG–TS, and sktime, we compare its runtime.

% Baseline Methods: It's often useful to include baseline methods or models in the benchmark to provide a reference point for performance comparison. - sktime as baseline 

\section{Results and Analysis}
%Analysis of the experimental results obtained from applying the AutoML tool.Comparison of the performance of the tool with other methods or benchmarks.Interpretation of the results and identification of trends, strengths, and weaknesses.

%results and evaluation: present the new theorems, designs, data, worked examples, and other results and interpret them (dito)

In this section, we will analyze the results of the AutoGluon framework, followed by the sktime methods, and then compare both based on different frequencies and domains. We will also assess whether tuning sktime methods results in improved performance and whether these results can compete with those achieved by AutoGluon. Tables \ref{table:4} through \ref{table:7} in the appendix summarize the results of the experiments with default settings for both frameworks. In these tables, \say{N/A} denotes model failures, indicating period constraints or errors in handling non-positive data during training; and \say{Timeout} represents cases where the estimator exceeded the 3600-second time limit and could not generate a forecast.
 \\ \textit{Note:} Despite attempting to load the Kaggle daily dataset, which consists of 145,063 daily time series representing web traffic for various Wikipedia pages, and the corresponding weekly dataset, which aggregates the same time series into weekly intervals for evaluation, both AutoGluon and sktime methods were unsuccessful in loading the datasets and stalled even after a 24-hour time frame.

\subsection{AUTOGLUON-TIMESERIES}
We first analyze the performance of the AutoGluon framework, and which methods it selects in detail. AutoGluon was initially evaluated using a 3600-second time budget and subsequently, its effectiveness with smaller time budgets such as a 600-second was evaluated. Throughout the experimentation, it was observed that the AutoGluon framework exhibited relatively consistent performance across varying training durations indicating longer training times do not necessarily yield a significant improvement in accuracy. However, there were a few instances where AutoGluon-600 (AG-TS 600) encountered difficulties in generating predictions within the 600-second time constraint, particularly with datasets like M4 Monthly (with 48,000 monthly time series), London Smart meters (with 5,560 half-hourly time series), Wind Farms (with lengthy minutely time series), and Temperature Rain (with 32,072 daily time series). This was primarily due to insufficient time to fit suitable models during the training process given the datasets' size and complexity. 
\begin{table}[!ht]
    \centering
    \begin{tabular}{c|c|c}
        \textbf{Metrics} & \textbf{AG-TS 600} & \textbf{AG-TS 3600} \\ \hline
        SMAPE & 0.33 & \textbf{0.31} \\ 
        %RMSE & 1.13E+17 & 1.11E+17 \\ 
        MASE & \textbf{24.23} & 30.10 \\ 
    \end{tabular}
\end{table}

On average, AutoGluon-3600 (AG-TS 3600) demonstrated slightly better performance than AutoGluon-600 for SMAPE, while for MASE, AutoGluon-600 outperformed AutoGluon-3600, as indicated in the table. To statistically assess the performance, Paired t-tests and Wilcoxon Signed-Rank Tests were conducted for each error metric. In the t-test, the test statistic was significant at a p-value of 0.05 for SMAPE, suggesting that longer training time can enhance performance. However, for MASE, there was no significant difference between the two variations. Regarding the Wilcoxon Signed-Rank Test, the test statistic was significant at a p-value of 0.05 for both metrics, indicating that AutoGluon-600 exhibited better performance than AutoGluon-3600 in terms of MASE, while the reverse was true for SMAPE.

% only the SMAPE value was significant at a p-value of 0.05, indicating that AutoGluon-3600 outperformed AutoGluon-600. In MASE, both variations of AutoGluon performed nearly equally. However, for the Wilcoxon Signed-Rank Test, the values for both metrics were significant at a p-value of 0.05, signifying that AutoGluon-600 performed better than AutoGluon-3600 in terms of MASE, while the reverse was true for SMAPE. 

% Specifically, AutoGluon-3600 achieved an average SMAPE value of \textbf{0.31}, whereas AutoGluon-600 had a slightly higher average SMAPE of \textbf{0.33}. For RMSE, AutoGluon-3600 recorded an average value of \textbf{1.11E+17}, which was slightly better than the \textbf{1.13E+17} average RMSE of AutoGluon-600. However, when assessing the Mean Absolute Scaled Error (MASE) metric, AutoGluon-600 outperformed AutoGluon-3600, with an average MASE value of \textbf{24.23} for AutoGluon-600 compared to \textbf{30.10} for AutoGluon-3600.

%In the other two metrics, both variations of AutoGluon performed nearly equally. However, for the Wilcoxon Signed-Rank Test, the values for all three metrics were significant at a p-value of 0.05, signifying that AutoGluon-600 performed better than AutoGluon-3600 in terms of MASE, while the reverse was true for SMAPE and RMSE.
%for both tests did not show significance, implying that the performance was similar between the two variations.
\begin{figure*}[th]
  \centering
  $\begin{array}{cc}
  \includegraphics[width=0.5\linewidth]{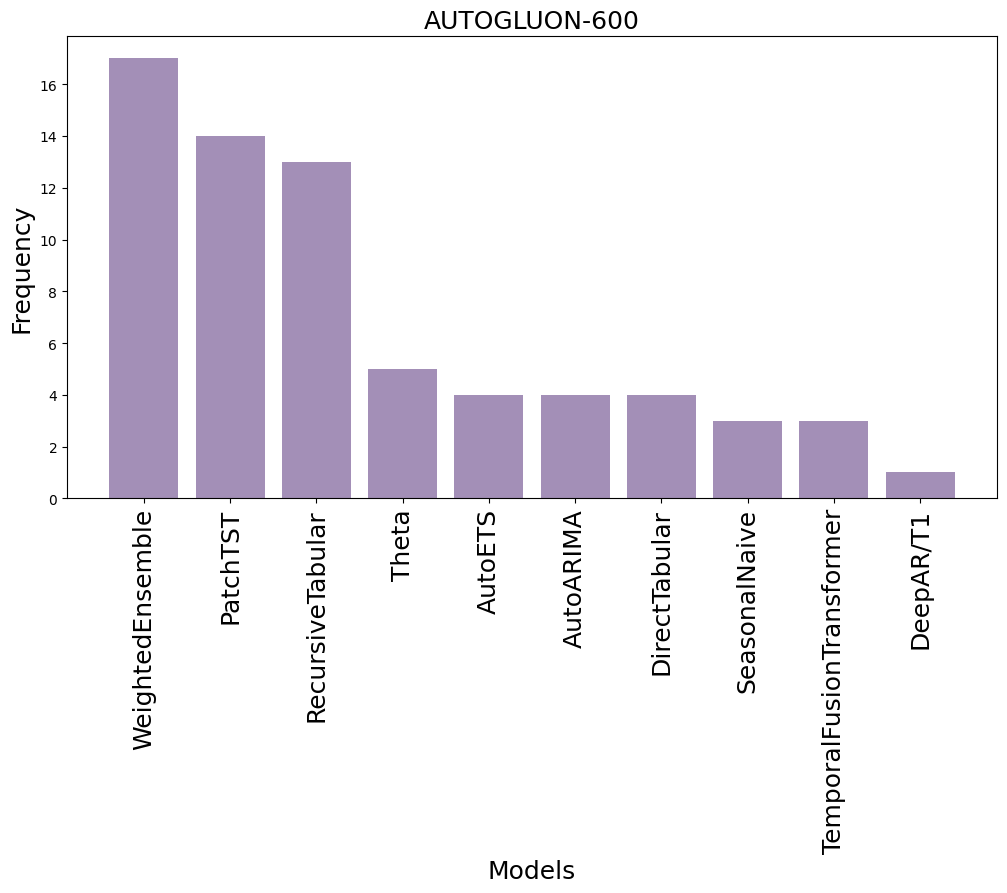} &
  \includegraphics[width=0.5\linewidth]{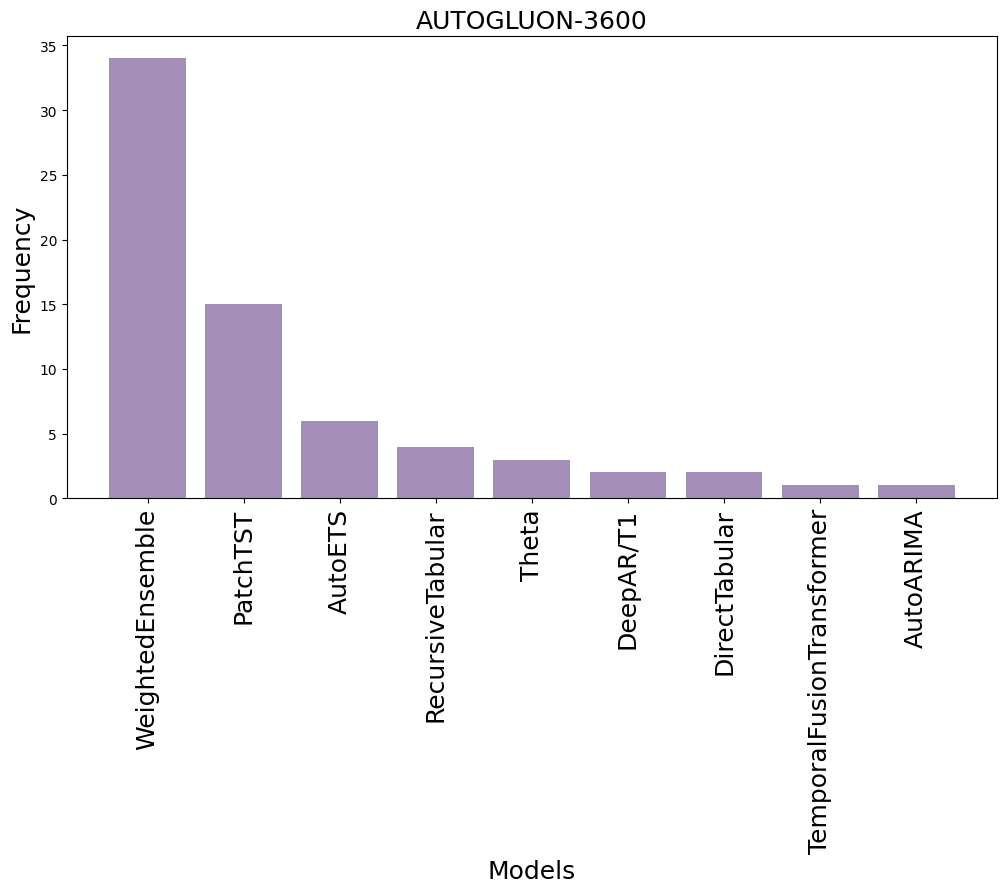} \\
  \mbox{(a)} & \mbox{(b)} 
  \end{array}$
  \caption{\label{Views:Figure} Frequency of different models trained by AutoGluon}
  \label{fig:ke}
\end{figure*}

\begin{table}[!ht]
    \centering
    \begin{tabular}{c|c|c}
        \textbf{Datasets} & \textbf{SMAPE} &  \textbf{MASE} \\ \hline
        London Smart Meters & RecursiveTabular &  TemporalFusionTransformer \\ 
        Wind Farms & RecursiveTabular &  SeasonalNaive \\ \hline
        Sunspot Daily & AutoARIMA & AutoARIMA   \\
        Saugeen River Flow & RecursiveTabular & WeightedEnsemble \\
        US Births & RecursiveTabular & RecursiveTabular \\
        Wind Power & Theta & Theta \\ \hline
    \end{tabular}    
\end{table}

To explore the prevalent methods selected within the AutoGluon framework, the models were categorized for both variations as illustrated in \textit{Figure} \ref{fig:ke}. The preferred model choices often leaned towards WeightedEnsemble,  RecursiveTabular, and PatchTST. Even for large datasets like London Smart Meters and Wind Farms,
and single very long-time series datasets such as Sunspot Daily, Saugeen River Flow, US Births, and Wind Power, RecursiveTabular was one of the preferred models, probably due to its efficient data processing methodology and effective handling of large and long time series without incurring excessive training times. Similarly, simple statistical models like AutoARIMA and Theta also performed well on single long time series due to their suitability for capturing patterns and trends within individual time series as compared to multiple time series within the dataset, where global models like PatchTST took precedence. WeightedEnsemble, on the other hand, consistently outperformed other methods, indicating a significant improvement in overall model performance achieved through the combination of predictions from multiple models.

\subsection{SKTIME}

% \begin{figure}[h!]
%     \centering
%     \includegraphics[width=0.74\textwidth]{rmsecount.png}
%     \caption{Wins and losses for RMSE}
%     \label{fig:rmsedomain}
% \end{figure}

% \begin{figure}[h!]
%     \centering
%     \includegraphics[width=0.74\textwidth]{smapecount.png}
%     \caption{Wins and losses for SMAPE}
%     \label{fig:smapedomain}
% \end{figure}

% \begin{figure}[h!]
%     \centering
%     \includegraphics[width=0.74\textwidth]{masecount.png}
%     \caption{Wins and losses for MASE}
%     \label{fig:masedomain}
% \end{figure}

\begin{figure*}[th]
  \centering
  $\begin{array}{cc}
  \includegraphics[width=0.5\linewidth]{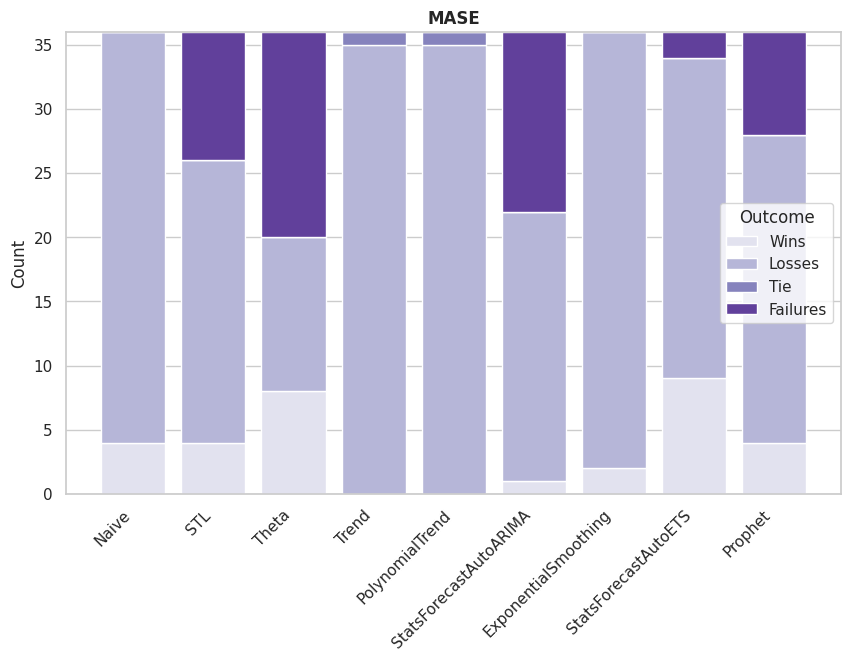} &
  \includegraphics[width=0.5\linewidth]{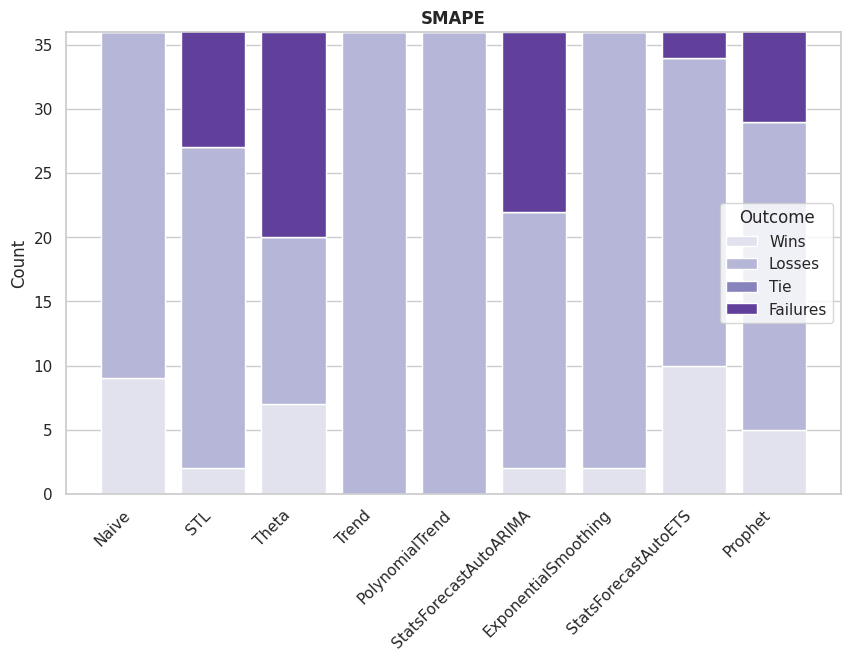} \\
  % \mbox{(a)} & \mbox{(b)} 
  \end{array}$
   \caption{\label{Views:metriccount} Wins and losses for all the sktime methods}
\end{figure*}

The sktime methods showed good performance overall with various models such as Naive, Trend, PolynomialTrend, Exponential smoothing, and StatsForecastAutoETS successfully producing forecasts for most datasets. However, there were some exceptions. The Wind power and the London Smart meters datasets posed challenges, as none of the models were able to produce forecasts for them. In addition, certain models like STL were unable to generate forecasts for datasets with a period of 1 (e.g., yearly) due to period $\geq 2$ constraint, and the Theta model had difficulty with datasets containing zero and negative values. Finally, models such as StatsForecastAutoARIMA and Prophet struggled to produce forecasts within the time limit of 3600 seconds for some datasets. Even when the time limit was extended from 1 hour to 4 hours, these methods still struggled to provide results for most of the datasets. Despite these errors and timeouts, the results of the study remain robust. The overview of wins and losses for each method over 36 datasets is given in Figure \ref{Views:metriccount}. We see that no single method outperforms others consistently; rather, their performance varies across different datasets, with MASE being more distinctive as compared to SMAPE.

% \begin{figure*}[th]
%   \centering
%   $\begin{array}{cc}
%   \includegraphics[width=0.5\linewidth]{rmsecount.png} &
%   \includegraphics[width=0.5\linewidth]{smapecount.png} \\
%   % \mbox{(a)} & \mbox{(b)} 
%   \end{array}$
%   % \caption{\label{Views:Figure} Wins and losses for (a) RMSE  \ (b) SMAPE \ (c) MASE}
% \end{figure*}

% \begin{figure}[h!]
%     \centering
%     \includegraphics[width=0.5\textwidth]{masecount.png}
%     \caption{Wins and losses for all the sktime methods}
%     \label{fig:metriccount}
% \end{figure}

%Some of the frameworks failed to generate forecasts on certain datasets. AutoARIMA, AutoTheta and StatEnsemble did not finish training on some datasets (Electricity–Hourly, KDD Cup 2018, and Pedestrian Counts) within 6 hours. This is caused by the poor scaling of these models to very long time series. DeepAR model fails on one dataset (Web Traffic Weekly) due to numerical errors encountered during training.

% On average, Naive has 0.350197425 for smape, 

% Trend has 0.523244064 for smape, 

% PolynomialTrend has 0.523245795 for smape, 

% Exponential smoothing has 0.442456374 for smape,

% StatsForecastAutoETS has 0.445345709 for smape,

% 9.17822E+12	9.17822E+12	9.17822E+12	9.17822E+12	9.71812E+12
% for mase

% 5.44E+16	2.51E+16	2.51E+16	4.22E+16	3.56E+16
% for rmse

\subsection{AUTOGLUON-TIMESERIES AND SKTIME}

To statistically compare the performance of all the methods on the various datasets from both frameworks, a critical difference (CD) diagram, and corresponding violin plots are presented. For CD diagrams, a non-parametric Friedman test is conducted where the algorithms are ranked for each dataset individually, and the best-performing algorithm receives a rank of 1, the second-best a rank of 2, and so on \citep{demvsar2006statistical}, to determine which pairs of algorithms exhibit significant differences. Subsequently, a post-hoc test, such as \say{Holm} \citep{pereira2015overview}, is applied to further assess the results. As for the violin plots, accuracy measures are rescaled to a range between 0 (indicating lower error) and 1 (indicating higher error) for each dataset. STL, Theta, StatsForecastAutoARIMA, and Prophet have been omitted from this comparison for the reasons mentioned above. Additionally, only datasets with available results, excluding M4 Monthly, Wind Power, London Smart Meters, Wind Farms, and Temperature Rain, totaling 33 in number, are taken into consideration. 

%Overall Pattern: Assess the overall pattern of the violins. A method with a narrower and taller violin might have a more concentrated range of SMAPE values, while a wider and flatter violin could indicate greater variability.
%Group Comparison: If you have multiple violins (one for each time series method) side by side, you can directly compare their shapes. Differences in width or height at specific points along the violins indicate variations in SMAPE values between the methods.
%Density and Distribution: Violin plots combine elements of box plots and kernel density plots. The width of the "violin" represents the density of the data at different points. A wider section indicates higher data density, giving you an idea of where the SMAPE values are more concentrated.

%For RMSE, the AUTOGLUON framework, and the StatsForcastAutoETS method show statistically significant differences compared to other methods. 

%It is also notable that the Autogluon framework performs better with sMAPE and MASE than with RMSE, a pattern evident in the violin plots as well.

\begin{figure}
\begin{minipage}{0.48\textwidth}
\centering
\resizebox{0.99\columnwidth}{!}{%
\begin{tikzpicture}[
  treatment line/.style={rounded corners=1.5pt, line cap=round, shorten >=1pt},
  treatment label/.style={font=\small},
  group line/.style={ultra thick},
]

\begin{axis}[
  clip={false},
  axis x line={center},
  axis y line={none},
  axis line style={-},
  xmin={1},
  ymax={0},
  scale only axis={true},
  width={\axisdefaultwidth},
  ticklabel style={anchor=south, yshift=1.3*\pgfkeysvalueof{/pgfplots/major tick length}, font=\small},
  every tick/.style={draw=black},
  major tick style={yshift=.5*\pgfkeysvalueof{/pgfplots/major tick length}},
  minor tick style={yshift=.5*\pgfkeysvalueof{/pgfplots/minor tick length}},
  title style={yshift=\baselineskip},
  xmax={9},
  ymin={-5.5},
  height={6\baselineskip},
  xtick={1,3,5,7,9},
  minor x tick num={1},
]

\draw[treatment line] ([yshift=-2pt] axis cs:1.8333333333333333, 0) |- (axis cs:1.0833333333333333, -2.5)
  node[treatment label, anchor=east] {AUTOGLUON-3600};
\draw[treatment line] ([yshift=-2pt] axis cs:2.9242424242424243, 0) |- (axis cs:1.0833333333333333, -3.5)
  node[treatment label, anchor=east] {AUTOGLUON - 600};
\draw[treatment line] ([yshift=-2pt] axis cs:3.6363636363636362, 0) |- (axis cs:1.0833333333333333, -4.5)
  node[treatment label, anchor=east] {PatchTST  };
\draw[treatment line] ([yshift=-2pt] axis cs:5.106060606060606, 0) |- (axis cs:1.0833333333333333, -5.5)
  node[treatment label, anchor=east] {StatsForecastAutoETS};
\draw[treatment line] ([yshift=-2pt] axis cs:5.242424242424242, 0) |- (axis cs:8.068181818181818, -6.0)
  node[treatment label, anchor=west] {ExponentialSmoothing};
\draw[treatment line] ([yshift=-2pt] axis cs:5.333333333333333, 0) |- (axis cs:8.068181818181818, -5.0)
  node[treatment label, anchor=west] {Naïve};
\draw[treatment line] ([yshift=-2pt] axis cs:6.287878787878788, 0) |- (axis cs:8.068181818181818, -4.0)
  node[treatment label, anchor=west] {RecursiveTabular};
\draw[treatment line] ([yshift=-2pt] axis cs:7.318181818181818, 0) |- (axis cs:8.068181818181818, -3.0)
  node[treatment label, anchor=west] {Trend};
\draw[treatment line] ([yshift=-2pt] axis cs:7.318181818181818, 0) |- (axis cs:8.068181818181818, -2.0)
  node[treatment label, anchor=west] {PolynomialTrend};
\draw[group line] (axis cs:2.9242424242424243, -2.3333333333333335) -- (axis cs:5.333333333333333, -2.3333333333333335);
\draw[group line] (axis cs:1.8333333333333333, -1.6666666666666667) -- (axis cs:3.6363636363636362, -1.6666666666666667);
\draw[group line] (axis cs:3.6363636363636362, -2.6666666666666665) -- (axis cs:6.287878787878788, -2.6666666666666665);
\draw[group line] (axis cs:6.287878787878788, -1.3333333333333333) -- (axis cs:7.318181818181818, -1.3333333333333333);

\end{axis}
\end{tikzpicture}
}
\end{minipage}
\hspace{0.01\textwidth}
\begin{minipage}{0.49\textwidth}
    \centering
    \includegraphics[width=0.99\textwidth]{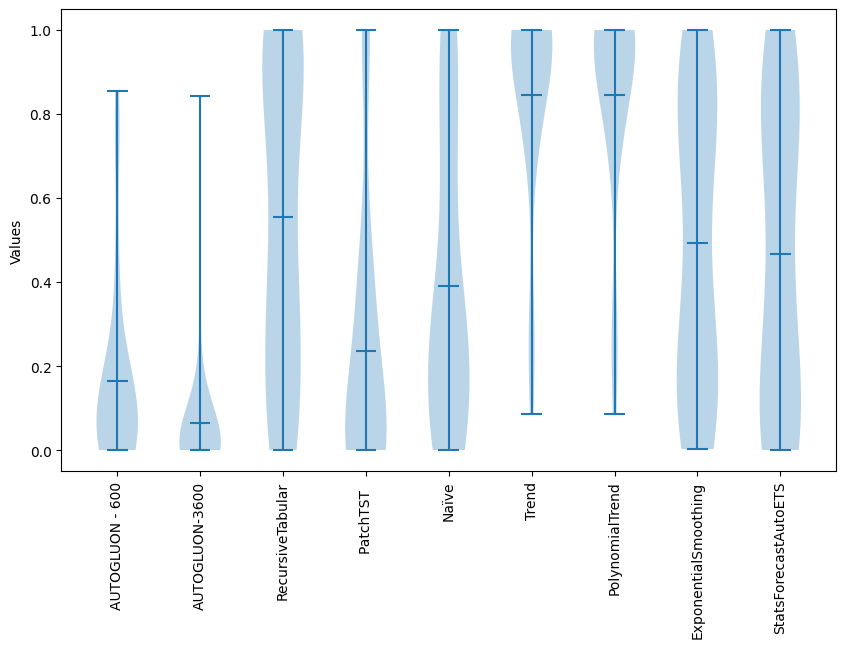}
\end{minipage}
\caption{CD diagram and violin plot for SMAPE}
\label{fig:smape}
\end{figure}

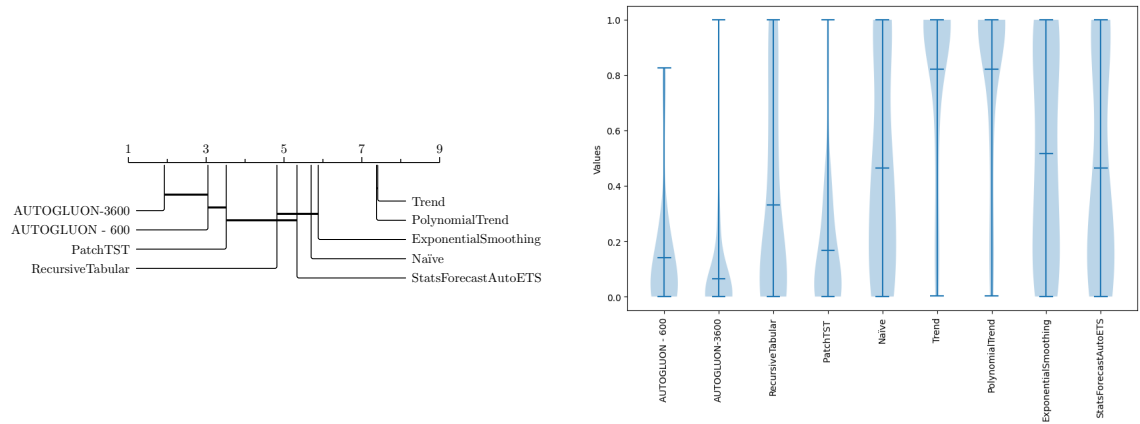
\begin{figure}
\begin{minipage}{0.48\textwidth}
\centering
\resizebox{0.99\columnwidth}{!}{%
\begin{tikzpicture}[
  treatment line/.style={rounded corners=1.5pt, line cap=round, shorten >=1pt},
  treatment label/.style={font=\small},
  group line/.style={ultra thick},
]

\begin{axis}[
  clip={false},
  axis x line={center},
  axis y line={none},
  axis line style={-},
  xmin={1},
  ymax={0},
  scale only axis={true},
  width={\axisdefaultwidth},
  ticklabel style={anchor=south, yshift=1.3*\pgfkeysvalueof{/pgfplots/major tick length}, font=\small},
  every tick/.style={draw=black},
  major tick style={yshift=.5*\pgfkeysvalueof{/pgfplots/major tick length}},
  minor tick style={yshift=.5*\pgfkeysvalueof{/pgfplots/minor tick length}},
  title style={yshift=\baselineskip},
  xmax={9},
  ymin={-5.5},
  height={6\baselineskip},
  xtick={1,3,5,7,9},
  minor x tick num={1},
]

\draw[treatment line] ([yshift=-2pt] axis cs:1.9242424242424243, 0) |- (axis cs:1.1742424242424243, -2.5)
  node[treatment label, anchor=east] {AUTOGLUON-3600};
\draw[treatment line] ([yshift=-2pt] axis cs:3.0454545454545454, 0) |- (axis cs:1.1742424242424243, -3.5)
  node[treatment label, anchor=east] {AUTOGLUON - 600};
\draw[treatment line] ([yshift=-2pt] axis cs:3.515151515151515, 0) |- (axis cs:1.1742424242424243, -4.5)
  node[treatment label, anchor=east] {PatchTST };
\draw[treatment line] ([yshift=-2pt] axis cs:4.818181818181818, 0) |- (axis cs:1.1742424242424243, -5.5)
  node[treatment label, anchor=east] {RecursiveTabular};
\draw[treatment line] ([yshift=-2pt] axis cs:5.333333333333333, 0) |- (axis cs:8.15909090909091, -6.0)
  node[treatment label, anchor=west] {StatsForecastAutoETS};
\draw[treatment line] ([yshift=-2pt] axis cs:5.696969696969697, 0) |- (axis cs:8.15909090909091, -5.0)
  node[treatment label, anchor=west] {Naïve};
\draw[treatment line] ([yshift=-2pt] axis cs:5.878787878787879, 0) |- (axis cs:8.15909090909091, -4.0)
  node[treatment label, anchor=west] {ExponentialSmoothing};
\draw[treatment line] ([yshift=-2pt] axis cs:7.378787878787879, 0) |- (axis cs:8.15909090909091, -3.0)
  node[treatment label, anchor=west] {PolynomialTrend};
\draw[treatment line] ([yshift=-2pt] axis cs:7.409090909090909, 0) |- (axis cs:8.15909090909091, -2.0)
  node[treatment label, anchor=west] {Trend};
\draw[group line] (axis cs:1.9242424242424243, -1.6666666666666667) -- (axis cs:3.0454545454545454, -1.6666666666666667);
\draw[group line] (axis cs:3.0454545454545454, -2.3333333333333335) -- (axis cs:3.515151515151515, -2.3333333333333335);
\draw[group line] (axis cs:3.515151515151515, -3.0) -- (axis cs:5.333333333333333, -3.0);
\draw[group line] (axis cs:4.818181818181818, -2.6666666666666665) -- (axis cs:5.878787878787879, -2.6666666666666665);
\draw[group line] (axis cs:7.378787878787879, -1.3333333333333333) -- (axis cs:7.409090909090909, -1.3333333333333333);

\end{axis}
\end{tikzpicture}
}
\end{minipage}
\hspace{0.01\textwidth}
\begin{minipage}{0.49\textwidth}
    \centering
    \includegraphics[width=0.99\textwidth]{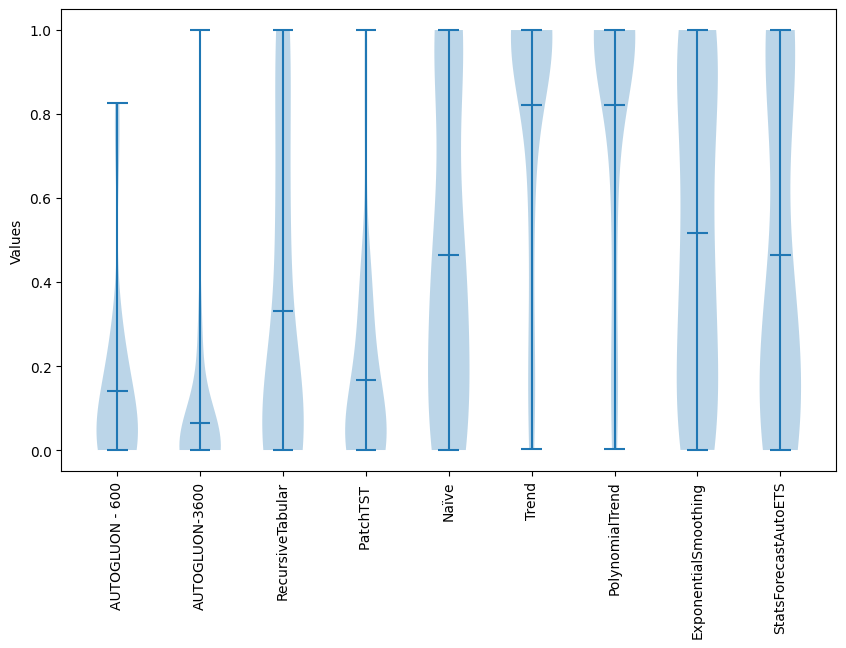}
\end{minipage}
\caption{CD diagram and violin plot for MASE}
\label{fig:mase}
\end{figure}

\textit{Figures} \ref{fig:smape}, and \ref{fig:mase} display the CD diagrams and associated violin plots for SMAPE, and MASE metrics respectively. The CD diagrams provide a visual representation of the average ranks assigned to each selected method while the violin plots display relative performance distribution. In the CD diagrams, lower ranks indicate better performance, and these ranks are averaged across all the results from the datasets. The connected algorithms indicate a lack of definitive distinction between each other, although they remain distinct from other algorithms in the comparison. For both SMAPE and MASE, the AutoGluon framework and Trend methods exhibit statistically significant differences compared to other methods. Examining the corresponding violin plots, we see that SMAPE and MASE values are concentrated near 0 for AutoGluon and around 1 for Trend methods, indicating better accuracy and lower errors for AutoGluon. Additionally, PatchTST, when trained individually within AutoGluon (results in table \ref{table:alone} of the appendix), demonstrates particularly favorable results, as seen in the CD and violin plots. This suggests that the state-of-the-art PatchTST approach excels in capturing and predicting time series patterns, showcasing its effectiveness in handling complex temporal patterns and adapting to diverse datasets, making it a robust choice in time series forecasting tasks.
\subsubsection{Frequency}
%https://www.aptech.com/blog/introduction-to-the-fundamentals-of-time-series-data-and-analysis/
To further assess the performance of these frameworks, we grouped the datasets based on their frequency, which denotes the intervals at which observations were recorded. This includes yearly (4 datasets), quarterly (4 datasets), monthly (6 datasets), weekly (5 datasets), daily (8 datasets), and hourly (6 datasets) as shown in tables \ref{smapefreq}, and \ref{masefreq}. For SMAPE, we see that AutoGluon emerges as the leading performer across all frequencies, except for monthly datasets where the naive approach takes precedence. Despite trend and polynomial methods having slightly higher average errors, they showcase lower variability, indicating reduced uncertainty in predictions for yearly and daily datasets. As for MASE, AutoGluon consistently outperforms other methods providing generalizable results in the majority of cases, with one exception for weekly where StatsForcastAutoETS takes the lead.

\begin{table}[!ht]
    \centering
    \begin{tabular}{|p{1.6cm}|p{0.2cm}|p{1.3cm}|p{1.3cm}|p{1.3cm}|p{1.25cm}|p{1.25cm}|p{1.82cm}|p{1.55cm}|}
    \hline
    \multirow{2}{*}{Frequency} & \multirow{2}{*}{~} & \multicolumn{7}{c|}{Forecasting Methods} \\ \cline{3-9}
    & & AG-TS 600 & AG-TS 3600 & Naive & Trend & Poly. Trend & Exp. Smoothing & AutoETS \\ \hline
    \multirow{2}{*}{Yearly} & $\overline{e}$ & 0.1888 & \textbf{0.1752} & 0.2471 & 0.2374 & 0.2374 & 0.2321 & 0.2103 \\ \cline{2-9}
    & $\sigma$ & 0.0650 & 0.0447 & 0.1192 & 0.0349 & 0.0349 & 0.0872 & 0.0942 \\ \hline
        \multirow{2}{*}{Quarterly} & $\overline{e}$ & 0.1508 & \textbf{0.1326} & 0.1478 & 0.2083 & 0.2083 & 0.1683 & 0.1713 \\ \cline{2-9}
         & $\sigma$ & 0.0573 & 0.0400 & 0.0363 & 0.0758 & 0.0758 & 0.0778 & 0.0814 \\ \hline
        \multirow{2}{*}{Monthly} & $\overline{e}$ & 0.4555 & 0.4257 & \textbf{0.2565} & 0.5093 & 0.5093 & 0.4578 & 0.4641 \\ \cline{2-9}
         & $\sigma$ & 0.6593 & 0.6615 & 0.1802 & 0.6346 & 0.6346 & 0.6441 & 0.6473 \\ \hline
        \multirow{2}{*}{Weekly} & $\overline{e}$ & 0.1225 & \textbf{0.1219} & 0.1649 & 0.2271 & 0.2271 & 0.1458 & 0.1360 \\ \cline{2-9}
         & $\sigma$ & 0.0492 & 0.0511 & 0.0932 & 0.1244 & 0.1244 & 0.0601 & 0.0428 \\ \hline
        \multirow{2}{*}{Daily} & $\overline{e}$ & 0.4079 & \textbf{0.4036} & 0.4669 & 0.5510 & 0.5510 & 0.4455 & 0.4410 \\ \cline{2-9}
         & $\sigma$ & 0.6290 & 0.6295 & 0.6439 & 0.5982 & 0.5982 & 0.6254 & 0.6295 \\ \hline
        \multirow{2}{*}{Hourly} & $\overline{e}$ & 0.4965 & \textbf{0.4128} & 0.5621 & 0.9023 & 0.9023 & 0.8514 & 0.8515 \\ \cline{2-9}
         & $\sigma$ & 0.2654 & 0.2250 & 0.5021 & 0.6440 & 0.6440 & 0.4912 & 0.4920 \\ \hline
    \end{tabular}
    \caption{Average SMAPE error ($\overline{e}$) and standard deviation ($\sigma$) for different methods grouped by frequency. The best average value (the lower, the better) is highlighted in bold.}
    \label{smapefreq}
\end{table}

\begin{table}[!ht]
    \centering
    \begin{tabular}{|p{1.6cm}|p{0.2cm}|p{1.25cm}|p{1.28cm}|p{1.5cm}|p{1.48cm}|p{1.48cm}|p{1.81cm}|p{1.6cm}|}
    \hline
    \multirow{2}{*}{Frequency} & \multirow{2}{*}{~} & \multicolumn{7}{c|}{Forecasting Methods} \\ \cline{3-9}
    & & AG-TS 600 & AG-TS 3600 & Naive & Trend & Poly. Trend & Exp. Smoothing & AutoETS \\ \hline
          \multirow{2}{*}{Yearly} & $\overline{e}$ & 3.1039 & \textbf{3.0357} & 3.8981 & 4.1568 & 4.1568 & 3.8538 & 3.2288 \\ \cline{2-9}
         & $\sigma$ & 0.3608 & 0.3888 & 0.7407 & 0.7356 & 0.7356 & 0.8026 & 0.5252 \\ \hline
        \multirow{2}{*}{Quarterly} & $\overline{e}$ & 1.5238 & \textbf{1.4447} & 1.7010 & 2.4873 & 2.4873 & 1.9908 & 1.9481 \\ \cline{2-9}
         & $\sigma$ & 0.2903 & 0.2904 & 0.2754 & 0.8518 & 0.8518 & 0.8461 & 0.9226 \\ \hline
        \multirow{2}{*}{Monthly} & $\overline{e}$ & 1.0670 & \textbf{0.9601} & 3.7E+12 & 3.7E+12 & 3.7E+12 & 3.7E+12 & 3.7E+12 \\ \cline{2-9}
         & $\sigma$ & 0.4051 & 0.4048 & 9E+12 & 9E+12 & 9E+12 & 9E+12 & 9E+12 \\ \hline
        \multirow{2}{*}{Weekly} & $\overline{e}$ & 154.33 & 194.98 & 2.0773 & 6.0273 & 6.0273 & 1.9340 & \textbf{1.8605} \\ \cline{2-9}
         & $\sigma$ & 343.25 & 434.08 & 0.8139 & 8.8814 & 8.8814 & 0.9149 & 0.9349 \\ \hline
        \multirow{2}{*}{Daily} & $\overline{e}$ & 2.2749 & \textbf{2.0566} & 3.9E+13 & 3.9E+13 & 3.9E+13 & 3.9E+13 & 3.9E+13 \\ \cline{2-9}
         & $\sigma$ & 2.0189 & 1.7936 & 1.1E+14 & 1.1E+14 & 1.1E+14 & 1.1E+14 & 1.1E+14 \\ \hline
        \multirow{2}{*}{Hourly} & $\overline{e}$ & 1.4952 & \textbf{1.3819} & 1.7206 & 4.5893 & 4.5893 & 4.1191 & 4.1967 \\ \cline{2-9}
         & $\sigma$ & 0.5256 & 0.6949 & 1.2905 & 3.4731 & 3.4731 & 3.9292 & 4.0925 \\ \hline
    \end{tabular}
    \caption{Average MASE error ($\overline{e}$) and standard deviation ($\sigma$) for different methods grouped by frequency. The best average value (the lower, the better) is highlighted in bold.}
    \label{masefreq}
\end{table}

\subsubsection{Domain}
\begin{figure}[h!]
    \centering
    \includegraphics[width=0.74\textwidth]{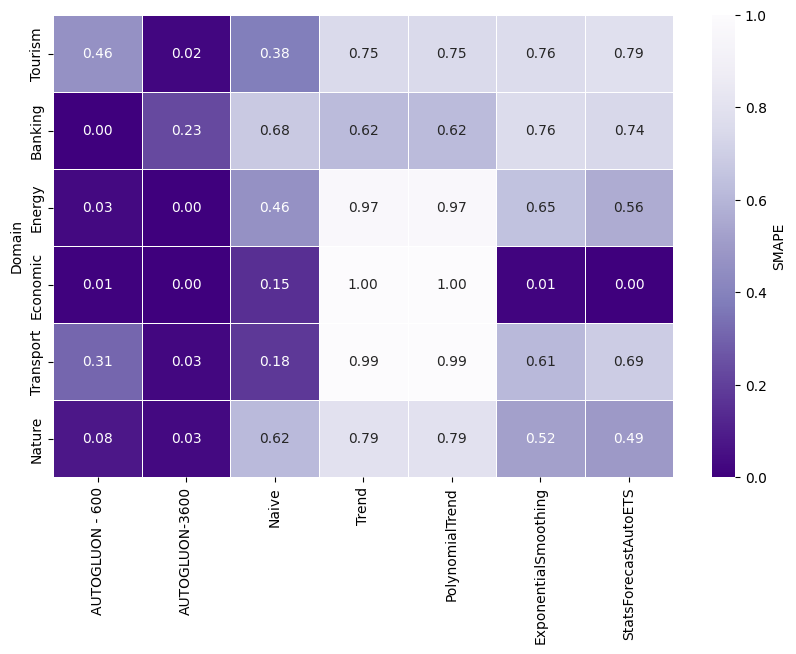}
    \caption{Average SMAPE error for different domains}
    \label{fig:smapedomain}
\end{figure}

\begin{figure}[h!]
    \centering
    \includegraphics[width=0.74\textwidth]{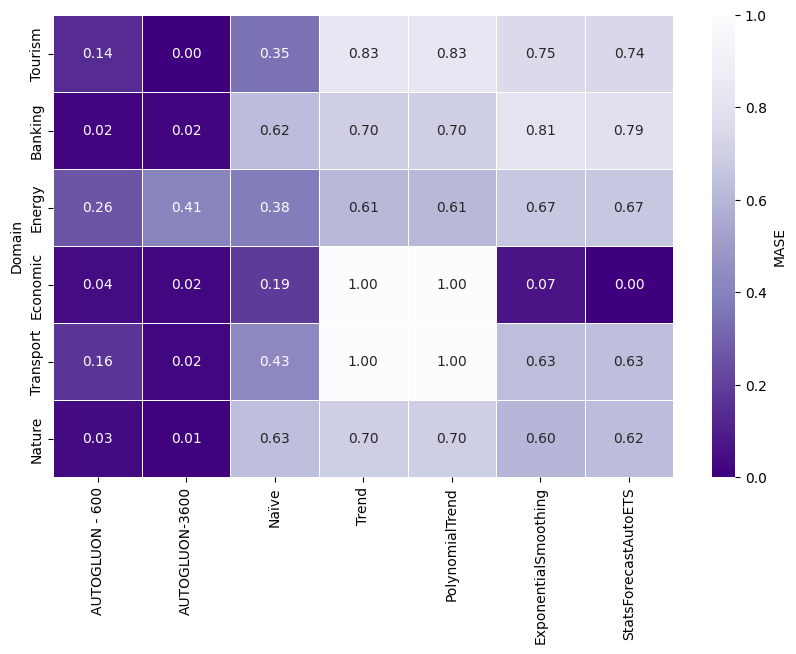}
    \caption{Average MASE error for different domains}
    \label{fig:masedomain}
\end{figure}

Similarly, to assess the variations in the performance of each method and to identify potential similarities between methods across different domains, we also grouped datasets based on domains, including Tourism, Banking, Energy, Economics, Transport, and Nature, totaling 20 datasets. The average errors of all the metrics are rescaled between 0 (best) and 1 (worst) for better visualization, as shown in figures \ref{fig:smapedomain}, and \ref{fig:masedomain}. AutoGluon consistently demonstrates good performance across diverse domains, for both SMAPE and MASE metrics under various time budgets attributed to its robustness and adaptability. In economic datasets, ExponentialSmoothing and StatsForcastAutoETS prove highly effective, although their performance wanes when applied to other domains. Conversely, Trend and Polynomial Trend fall short in transport, economic, and energy domains for SMAPE, as well as in transport and economic domains for MASE. Interestingly, Naive consistently outperforms other methods within the sktime framework, owing to its simplicity, making it less prone to overfitting or misinterpretation of patterns.

\subsection{TUNING}
For the majority of the datasets, we encountered evaluation challenges due to two primary reasons: first, the time limit of 4 hours for evaluation was exceeded, and second, a worker process, managed by the executor, was abruptly terminated without generating any forecast, probably due to excessive memory usage or segmentation fault. The individual scores for each metric are given in Tables \ref{smape_tuning}, and \ref{mase_tuning} of the appendix. The datasets that were abruptly terminated are excluded from this list. 

\begin{table}[!ht]
    \centering
    %\begin{tabular}{lccccccc}
    \begin{tabular}{p{1.3cm}p{1cm}p{1.6cm}p{1.6cm}p{1.6cm}p{1.6cm}p{1.7cm}p{1.7cm}}
    \toprule
        & & Naive & STL & Theta & Trend & Poly. & Exp. \\ 
        & & & & & & Trend & Smoothing \\ \midrule
        \multirow{2}{*}{SMAPE} & Before & \textbf{0.23} & 0.32 & 0.19 & 0.34 & 0.35 & \textbf{0.34} \\
        & After & 0.32 & \textbf{0.28} & 0.19 & \textbf{0.27} & \textbf{0.30} & 0.37 \\ \midrule
        \multirow{2}{*}{MASE} & Before & 1.44E+13 & 2.75E+13 & 8.55E+10 & 1.65E+13 & 1.70E+13 & 1.32E+13 \\
        & After & 1.44E+13 & 2.75E+13 & 8.55E+10 & 1.65E+13 & 1.74E+13 & 1.32E+13 \\ \bottomrule
    \end{tabular}
    \caption{Before and after tuning the sktime models}
    \label{tuningresults}
\end{table}

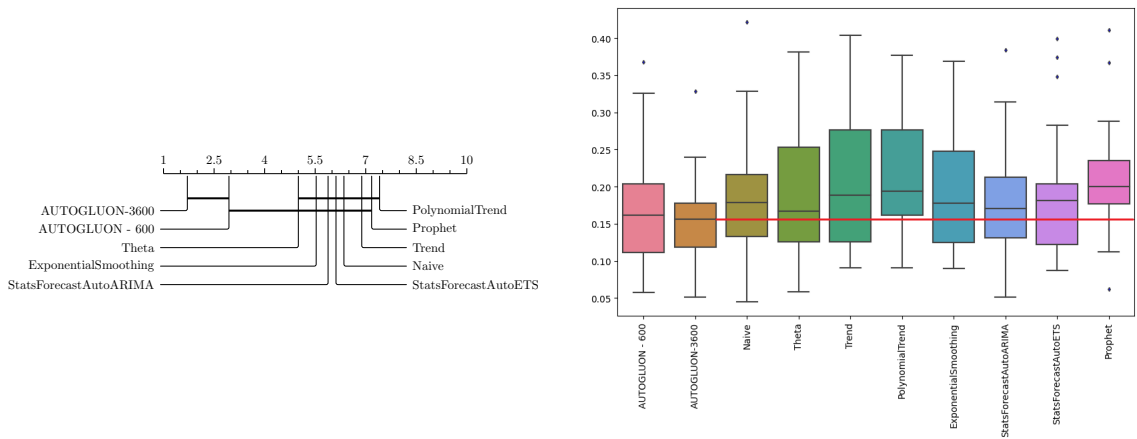
\begin{figure}
\begin{minipage}{0.48\textwidth}
\centering
\resizebox{0.99\columnwidth}{!}{%
\begin{tikzpicture}[
  treatment line/.style={rounded corners=1.5pt, line cap=round, shorten >=1pt},
  treatment label/.style={font=\small},
  group line/.style={ultra thick},
]

\begin{axis}[
  clip={false},
  axis x line={center},
  axis y line={none},
  axis line style={-},
  xmin={1},
  ymax={0},
  scale only axis={true},
  width={\axisdefaultwidth},
  ticklabel style={anchor=south, yshift=1.3*\pgfkeysvalueof{/pgfplots/major tick length}, font=\small},
  every tick/.style={draw=black},
  major tick style={yshift=.5*\pgfkeysvalueof{/pgfplots/major tick length}},
  minor tick style={yshift=.5*\pgfkeysvalueof{/pgfplots/minor tick length}},
  title style={yshift=\baselineskip},
  xmax={10},
  ymin={-6.5},
  height={7\baselineskip},
  xtick={1.0,2.5,4.0,5.5,7.0,8.5,10.0},
  minor x tick num={2},
]

\draw[treatment line] ([yshift=-2pt] axis cs:1.7058823529411764, 0) |- (axis cs:0.872549019607843, -2.0)
  node[treatment label, anchor=east] {AUTOGLUON-3600};
\draw[treatment line] ([yshift=-2pt] axis cs:2.9411764705882355, 0) |- (axis cs:0.872549019607843, -3.0)
  node[treatment label, anchor=east] {AUTOGLUON - 600};
\draw[treatment line] ([yshift=-2pt] axis cs:5.0, 0) |- (axis cs:0.872549019607843, -4.0)
  node[treatment label, anchor=east] {Theta};
\draw[treatment line] ([yshift=-2pt] axis cs:5.529411764705882, 0) |- (axis cs:0.872549019607843, -5.0)
  node[treatment label, anchor=east] {ExponentialSmoothing};
\draw[treatment line] ([yshift=-2pt] axis cs:5.882352941176471, 0) |- (axis cs:0.872549019607843, -6.0)
  node[treatment label, anchor=east] {StatsForecastAutoARIMA};
\draw[treatment line] ([yshift=-2pt] axis cs:6.117647058823529, 0) |- (axis cs:8.245098039215687, -6.0)
  node[treatment label, anchor=west] {StatsForecastAutoETS};
\draw[treatment line] ([yshift=-2pt] axis cs:6.352941176470588, 0) |- (axis cs:8.245098039215687, -5.0)
  node[treatment label, anchor=west] {Naive};
\draw[treatment line] ([yshift=-2pt] axis cs:6.882352941176471, 0) |- (axis cs:8.245098039215687, -4.0)
  node[treatment label, anchor=west] {Trend};
\draw[treatment line] ([yshift=-2pt] axis cs:7.176470588235294, 0) |- (axis cs:8.245098039215687, -3.0)
  node[treatment label, anchor=west] {Prophet};
\draw[treatment line] ([yshift=-2pt] axis cs:7.411764705882353, 0) |- (axis cs:8.245098039215687, -2.0)
  node[treatment label, anchor=west] {PolynomialTrend};
\draw[group line] (axis cs:1.7058823529411764, -1.3333333333333333) -- (axis cs:2.9411764705882355, -1.3333333333333333);
\draw[group line] (axis cs:2.9411764705882355, -2.0) -- (axis cs:7.176470588235294, -2.0);
\draw[group line] (axis cs:5.0, -1.3333333333333333) -- (axis cs:7.411764705882353, -1.3333333333333333);

\end{axis}
\end{tikzpicture}
}
\end{minipage}
\hspace{0.01\textwidth}
\begin{minipage}{0.49\textwidth}
    \centering
    \includegraphics[width=0.99\textwidth]{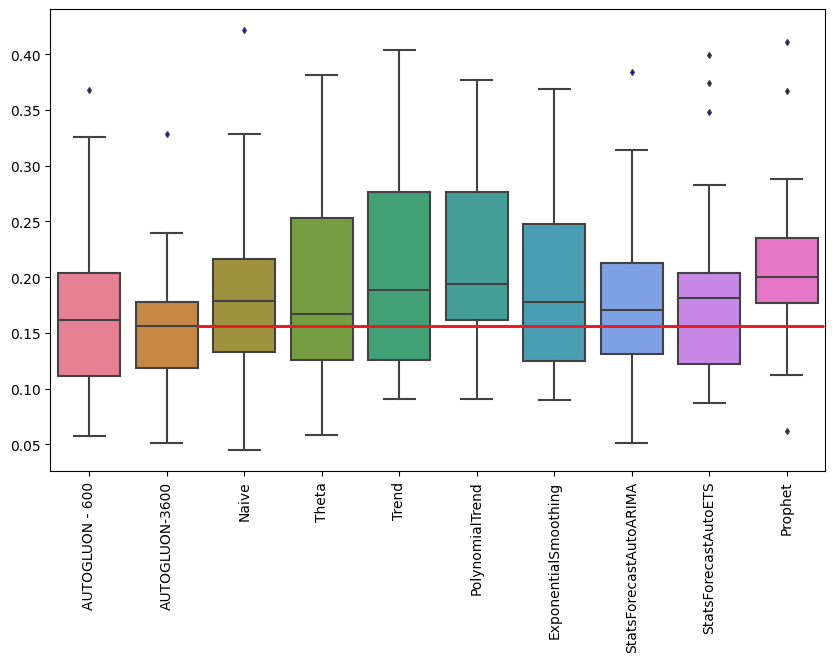}
\end{minipage}
\caption{CD diagram and box plot for SMAPE}
\label{fig:tuningsmape}
\end{figure}

From table \ref{tuningresults}, we see that, on average, the tuning did not significantly increase the performance of these methods when compared to default settings for MASE. As for SMAPE, tuning did show positive effects for methods such as STL, Trend, and Polynomial Trend. Theta, however, was able to produce a forecast for the datasets that previously yielded no results when deseasonalize was set to False. Since sktime models did not show improved results with tuning for MASE, further comparison with the AutoGluon framework was not pursued. However, for SMAPE, we conducted a comparison between AutoGluon and tuned versions of Trend, Polynomial Trend, and Theta, along with other sktime methods using their default settings across 17 datasets as shown in figure \ref{fig:tuningsmape}. STL was excluded from this comparison due to a reduced sample size of generated forecasts (limited to 9 datasets). In the CD diagram, we see that AutoGluon ranks the highest, followed by other sktime methods such as Theta, ExponentialSmoothing, and StatsForecastAutoARIMA. The corresponding box plot further reinforces this observation, depicting a lower median and variance for AutoGluon-3600 suggesting not only better performance but also greater consistency in forecasting accuracy across various datasets. Although methods like Theta and StatsForecastAutoARIMA, exhibit slightly higher medians and greater variance compared to AutoGluon, they perform relatively well overall with some variability.

\section{Conclusion and Future Work}
%Summary of the key findings and conclusions drawn from the research.Discussion on the implications of the research and its contributions to the field.Identification of potential avenues for future research and improvement in AutoML for time series.

%discussion: show why the results "solve" the problem and to what extent,  describe what your research or design adds to the field's knowledge, what remains to be done, and what other problems may be triggered; give hints for further study 
In summary, this study has explored whether AutoML frameworks outperform the state-of-the-art in time series forecasting, and which methods perform best overall when properly tuned through a comprehensive benchmarking approach. Two prominent frameworks, AutoGluon-Timeseries, and sktime, were thoroughly evaluated and analyzed for their applicability to diverse datasets from various domains. Overall, both variations of AutoGluon showed relatively better performance when tested across a range of datasets as compared to different methods in sktime. However, there were also a few instances where sktime methods, such as Naive and StatsForecastAutoETS, outperformed AutoGluon, particularly in monthly and weekly datasets. Notably, among other methods within the AutoGluon framework, PatchTST achieved commendable results, and the use of ensembling techniques showed improved performance. Furthermore, certain methods like StatsforcastARIMA and Prophet required significant time to generate forecasts for some datasets; other methods such as STL and Theta failed to produce forecasts for numerous datasets, and tuning did not significantly increase their performance. On the other hand, methods like Naive, Trend, PolynomialTrend, ExponentialSmoothing, and StatsForecastAutoETS performed well with default configurations. Although tuning had some impact on sktime algorithms Trend and PolynomialTrend, they still fell short of AutoGluon's performance.

%However, the performance of AutoGluon was not promising for RMSE which was also the case overall.

% In the future, the study could be extended to include more frameworks and datasets to increase adaptability and scalability. There is also a  potential to integrate these sktime methods with other tabular and deep learning models to develop automated machine learning (AutoML) tools streamlining the data preprocessing, feature engineering, hyperparameter optimization, method selection, and ensembling, significantly speeding up the process of selecting the most suitable approach for the given task. Additionally, we could also leverage the empirical data to construct a meta-model that can offer recommendations on which forecasting techniques are most suitable for specific datasets.

An extension of this study could involve the inclusion of additional frameworks, datasets, and global models to enhance adaptability and scalability. The empirical data can be further leveraged to facilitate the construction of a meta-model capable of providing recommendations regarding the most suitable forecasting techniques for specific datasets. The feasibility and benefits of applying transfer learning techniques within domains to improve performance on a different but related type can be explored. Lastly, the sktime framework can be used to integrate with other automated machine learning (AutoML) tools to streamline processes such as data preprocessing, feature engineering, hyperparameter optimization, method selection, and ensembling, thereby significantly speeding up the selection process for a given task and potentially improving the overall performance.
%https://www.proquest.com/docview/226915773?pq-origsite=gscholar&fromopenview=true - use this 

% Cross-Domain Generalization:

% Investigate the generalization capabilities of time series forecasting models across different domains. Understanding whether models trained in one domain can be effectively transferred to another could have practical implications, especially in situations where labeled data is scarce.

% Transfer Learning in Time Series Forecasting:

% Explore the feasibility and benefits of applying transfer learning techniques to time series forecasting. This could involve leveraging knowledge gained from forecasting one type of time series data to improve the performance on a different but related type.

%\newpage
% Acknowledgements should go at the end, before appendices and references

\acks{ Joaquin Vanschoren and Oleksandr Shchur acknowledge support from the European Research Council (ERC) under grant no. 952215 (TAILOR).}

%I wish to express my gratitude to my supervisor, Joaquin Vanschoren, and Oleksandr Shchur for their time and assistance provided throughout this project. I am also grateful to my family and friends for their unwavering support.}


\newpage
\appendix
\section*{Appendix A.}
\label{app:theorem}

% Note: in this sample, the section number is hard-coded in. Following
% proper LaTeX conventions, it should properly be coded as a reference:

%In this appendix we prove the following theorem from
%Section~\ref{sec:textree-generalization}:

\subsection*{DATASET DETAILS}
\begin{itemize}
    \item M1 Yearly: Contains 181 yearly time series used in the M1 forecasting competition. The series belongs to 7 different domains: macro 1, macro 2, micro 1, micro 2, micro 3, industry, and demographic.
    \item M1 Quarterly: Contains 203 quarterly time series used in the M1 forecasting competition. The series belongs to 7 different domains: macro 1, macro 2, micro 1, micro 2, micro 3, industry, and demographic.
    \item M1 Monthly: Contains 617 monthly time series used in the M1 forecasting competition. The series belongs to 7 different domains: macro 1, macro 2, micro 1, micro 2, micro 3, industry, and demographic.
    \item M3 Yearly: Contains 645 yearly time series used in the M3 forecasting competition. The series belongs to 6 different domains: demographic, micro, macro, industry, finance, and other.
    \item M3 Quarterly: Contains 756 quarterly time series used in the M3 forecasting competition. The series belongs to 5 different domains: demographic, micro, macro, industry, and finance.
    \item M3 Monthly: Contains 1428 monthly time series used in the M3 forecasting competition. The series belongs to 6 different domains: demographic, micro, macro, industry, finance, and other.
    \item M4 Yearly: Contains 23000 yearly time series used in the M4 forecasting competition.
    \item M4 Quarterly: Contains 24000 quarterly time series used in the M4 forecasting competition.
    \item M4 Monthly: Contains 48000 monthly time series used in the M4 forecasting competition. The series belongs to 6 different domains: demographic, micro, macro, industry, finance, and other.
    \item M4 Weekly: Contains 359 weekly time series used in the M4 forecasting competition.
    \item M4 Daily: Contains 4227 daily time series used in the M4 forecasting competition.
    \item M4 Hourly: Contains 414 hourly time series used in the M4 forecasting competition.
    \item Tourism Yearly: Contains 518 yearly time series used in the Kaggle Tourism forecasting competition.
    \item Tourism Quarterly: Contains 427 quarterly time series used in the Kaggle Tourism forecasting competition.
    \item Tourism Monthly: Contains 366 monthly time series used in the Kaggle Tourism forecasting competition.
    \item London Smart Meters without missing values: Contains 5560 half-hourly time series that represent the energy consumption readings of London households in kilowatt hour (kWh) from November 2011 to February 2014. The missing values are replaced by carrying forward the corresponding last observations (LOCF method).
    \item Wind Farms without missing values: Contains very long minute time series representing the wind power production of 339 wind farms in Australia. The missing values are replaced by zeros.
    \item Bitcoin without Missing Values: Contains the potential influencers of the bitcoin price with a total of 18 daily time series including hash rate, block size, mining difficulty, etc. The missing values are replaced by carrying forward the corresponding last-seen observations (LOCF method).
    \item Melbourne Pedestrian Counts: Contains hourly pedestrian counts captured from 66 sensors in Melbourne city starting from May 2009 up to 2020-04-30.
    \item Vehicle Trips without Missing Values: Contains 329 daily time series representing the number of trips and vehicles belonging to a set of for-hire vehicle (FHV) companies. The missing values are replaced by carrying forward the corresponding last-seen observations (LOCF method).
    \item KDD Cup without Missing Values: Contains 270 long hourly time series representing the air quality levels in 59 stations in 2 cities: Beijing (35 stations) and London (24 stations) from 01/01/2017 to 31/03/2018. The leading missing values of a given series are replaced by zeros and the remaining missing values are replaced by carrying forward the corresponding last observations (LOCF method). 
    \item NN5 Daily without Missing Values: Contains 111-time series from the banking domain. A missing value on a particular day is replaced by the median across all the same days of the week along the whole series.
    \item NN5 Weekly: Contains 111 weekly time series from the banking domain. A missing value on a particular day is replaced by the median across all the same days of the week along the whole series and then aggregated into weekly.
    \item Solar Weekly: Contains 137-time series representing the weekly solar power production in Alabama state in 2006. 
    \item Electricity Hourly: Contains 321 hourly time series of electricity consumption of 370 clients recorded in 15-minute periods in Kilowatt (kW) from 2012 to 2014.
    \item Electricity Weekly: Contains 321 weekly time series of electricity consumption of 370 clients recorded in 15-minute periods in Kilowatt (kW) from 2012 to 2014.
    \item Car Parts without Missing Values: Contains 2674 intermittent monthly time series that represent car parts sales from January 1998 to March 2002. The missing values are replaced by zeros.
    \item FRED-MD: Contains 107 monthly time series showing a set of macroeconomic indicators from the Federal Reserve Bank.
    \item Traffic Hourly: Contains 862 hourly time series showing the road occupancy rates on the San Francisco Bay area freeways from 2015 to 2016.
    \item Traffic Weekly: Contains 862 weekly time series showing the road occupancy rates on the San Francisco Bay area freeways from 2015 to 2016.
    \item Rideshare without Missing Values: Contains various hourly time series representations of attributes related to Uber and Lyft rideshare services for various locations in New York between 26/11/2018 and 18/12/2018. The missing values are replaced by zeros. 
    \item Hospital: Contains 767 monthly time series that represent the patient counts related to medical products from January 2000 to December 2006.
    \item COVID-19 Deaths: Contains 266 daily time series that represent the COVID-19 deaths in a set of countries and states from 22/01/2020 to 20/08/2020.
    \item Temperature Rain without Missing Values: Contains 32072 daily time series showing the temperature observations and rain forecasts, gathered by the Australian Bureau of Meteorology for 422 weather stations across Australia, between 02/05/2015 and 26/04/2017. The missing values are replaced by zeros.
    \item Sunspot Daily without Missing Values: Contains a single very long daily time series of sunspot numbers from 1818-01-08 to 2020-05-31. The missing values are replaced by carrying forward the corresponding last observations (LOCF method).
    \item Saugeen River Flow: Contains a single very long time series of length 23741 representing the daily mean flow of the Saugeen River at Walkerton in cubic meters per second from 01/01/1915 to 31/12/1979.
    \item US Births: Contains a single very long daily time series of length 7305 representing the number of births in the US from 01/01/1969 to 31/12/1988.
    \item Wind Power: Contains a single very long daily time series of length 7397147 representing the wind power production in MW recorded every 4 seconds starting from 01/08/2019.   
    \item Kaggle Wikipedia Web Traffic Daily with Missing Values: Contains 145063 daily time series representing the number of hits or web traffic for a set of Wikipedia pages from 2015-07-01 to 2017-09-10. The missing values are replaced by zeros.
    \item Kaggle Wikipedia Web Traffic Weekly: Contains 145063 time series representing the number of hits or web traffic for a set of Wikipedia pages from 2015-07-01 to 2017-09-05, after aggregating them into weekly. The missing values are replaced by zeros before aggregation.
\end{itemize}

\begin{table}[!ht]
    %\raggedleft
    \centering
    \begin{tabular}{|| l || r || r ||}
    \hline\hline
        \textbf{Dataset} & \textbf{Frequency} & \textbf{fh} \\ \hline\hline
        M1 yearly & Yearly & 6 \\ \hline
        M1 quaterly & Quaterly & 6 \\ \hline
        M1 monthly & Monthly & 18 \\ \hline
        M3 yearly & Yearly & 6 \\ \hline
        M3 quaterly & Quaterly & 8 \\ \hline
        M3 monthly & Monthly & 18 \\ \hline
        M4 Yearly & Yearly & 6 \\ \hline
        M4 Quarterly & Quarterly & 8 \\ \hline
        M4 Monthly  & Monthly & 18 \\ \hline
        M4 Weekly & Weekly & 13 \\ \hline
        M4 Daily & Daily & 14 \\ \hline
        M4 Hourly & Hourly & 48 \\ \hline
        Tourism Yearly & Yearly & 4 \\ \hline
        Tourism Quarterly & Quarterly & 8 \\ \hline
        Tourism Monthly  & Monthly & 24 \\ \hline
        Bitcoin Dataset without Missing Values & Daily & 30 \\ \hline
        Melbourne Pedestrian Counts & Hourly & 24 \\ \hline
        NN5 Daily Dataset without Missing Values & Daily & 56 \\ \hline
        NN5 Weekly & Weekly & 8 \\ \hline
        Solar Weekly & Weekly & 5 \\ \hline
        Electricity Hourly & Hourly & 168 \\ \hline
        Electricity Weekly & Weekly & 8 \\ \hline
        Car Parts (without Missing Values) & Monthly & 12 \\ \hline
        FRED-MD & Monthly & 12 \\ \hline
        Traffic Hourly & Hourly & 168 \\ \hline
        Traffic Weekly & Weekly & 8 \\ \hline
        Hospital  & Monthly & 12 \\ \hline
        COVID-19 Deaths & Daily & 30 \\ \hline
        Sunspot Daily without Missing Values & Daily & 30 \\ \hline
        Saugeen River Flow & Daily & 30 \\ \hline
        US Births & Daily & 30 \\ \hline
        Wind Power Dataset 4 Seconds Observations & 4\_seconds & 21600 \\ \hline
        Vehicle Trips without Missing Values & Daily & 30 \\ \hline
        Rideshare without Missing Values & Hourly & 168 \\ \hline
        Temperature Rain without Missing Values & Daily & 30 \\ \hline
        KDD Cup without Missing Values & Hourly & 168 \\ \hline
        London Smart Meters Dataset without Missing Values & Half-Hourly & 48 \\ \hline
        Wind Farms Dataset without Missing Values & Minutely & 1440 \\ \hline
        Kaggle Wikipedia Web Traffic Daily without Missing Values & Daily & 59 \\ \hline
        Kaggle Wikipedia Web Traffic Weekly & Weekly & 8 \\ \hline\hline
    \end{tabular}
\end{table}

\begin{table}[!ht]
    \centering
    \begin{tabular}{ |p{3.7cm}|p{1.8cm}|p{1.8cm}|p{1.8cm}|p{1.8cm}|p{1.8cm}|  }
    % \hline
    % \multicolumn{6}{|c|}{SMAPE} \\
    % \hline
    \hline
    \textbf{Dataset} & \textbf{AG-TS 600} & \textbf{AG-TS 3600} & \textbf{Naive} & \textbf{STL} & \textbf{Theta}\\ \hline
        M1 yearly & 0.1712 & 0.1698 & 0.2243 & N/A & 0.1922 \\ \hline
        M1 quarterly & 0.1619 & 0.1565 & 0.1894 & 0.1776 & 0.1568 \\ \hline
        M1 monthly & 0.1712 & 0.162 & 0.1731 & 0.1867 & 0.1622 \\ \hline
        M3 yearly & 0.1600 & 0.1540 & 0.1788 & N/A & 0.1673 \\ \hline
        M3 quarterly & 0.1018 & 0.0951 & 0.1107 & 0.0998 & 0.0914 \\ \hline
        M3 monthly & 0.1555 & 0.1494 & 0.1724 & 0.1671 & 0.1442 \\ \hline
        M4 Yearly & 0.1397 & 0.1377 & 0.1634 & N/A & 0.1436 \\ \hline
        M4 Quarterly & 0.1122 & 0.1023 & 0.1252 & 0.114 & 0.1034 \\ \hline
        M4 Monthly & Timeout & 0.1377 & 0.1599 & Timeout & 0.128 \\ \hline
        M4 Weekly & 0.0716 & 0.0648 & 0.0916 & N/A & 0.0911 \\ \hline
        M4 Daily & 0.0305 & 0.0303 & 0.0374 & 0.0364 & 0.0307 \\ \hline
        M4 Hourly & 0.1356 & 0.1318 & 0.1391 & 0.231 & 0.1814 \\ \hline
        Tourism Yearly & 0.2844 & 0.2394 & 0.4217 & N/A & 0.3343 \\ \hline
        Tourism Quarterly & 0.2273 & 0.1764 & 0.1661 & 0.1496 & N/A \\ \hline
        Tourism Monthly & 0.3254 & 0.1880 & 0.2167 & 0.1957 & N/A \\ \hline
        NN5 Daily & 0.2132 & 0.2239 & 0.2647 & 0.275 & N/A \\ \hline
        NN5 Weekly & 0.1094 & 0.1183 & 0.1327 & N/A & 0.1203 \\ \hline
        Solar Weekly & 0.2036 & 0.2036 & 0.328 & N/A & 0.2494 \\ \hline
        Traffic Weekly & 0.1243 & 0.12433 & 0.1307 & N/A & N/A \\ \hline
        Electricity Weekly & 0.1035 & 0.0994 & 0.1414 & N/A & N/A \\ \hline
        Bitcoin & 0.2058 & 0.2058 & 0.2206 & 0.3273 & 0.302 \\ \hline
        Melbourne Pedestrian Counts & 0.683 & 0.5094 & 0.5146 & 0.7413 & N/A \\ \hline
        Traffic Hourly & 0.6928 & 0.3963 & 0.3484 & 0.576 & N/A \\ \hline
        Electricity Hourly & 0.2052 & 0.199 & 0.2242 & 0.2366 & N/A \\ \hline
        Car Parts & 1.7929 & 1.7747 & 0.6204 & 1.8299 & N/A \\ \hline
        FRED-MD & 0.1063 & 0.1028 & 0.1459 & 0.1206 & N/A \\ \hline
        Hospital & 0.1818 & 0.1773 & 0.2103 & 0.2132 & 0.1805 \\ \hline
        COVID-19 Deaths & 0.111 & 0.111 & 0.1863 & 0.1645 & N/A \\ \hline
        Sunspot Daily & 1.9349 & 1.9349 & 1.9833 & 1.9802 & N/A \\ \hline
        US Births & 0.0579 & 0.0517 & 0.045 & 0.0513 & 0.0582 \\ \hline
        Saugeen River Flow & 0.3424 & 0.3424 & 0.6695 & 0.6151 & 0.3604 \\ \hline
        Wind Power & 0.4336 & 0.4336 & N/A & N/A & N/A \\ \hline
        London Smart Meters & Timeout & 0.6689 & N/A & N/A & N/A \\ \hline
        Wind Farms & Timeout & 1.3986 & 0.4305 & N/A & N/A \\ \hline
        Vehicle Trips  & 0.3675 & 0.3287 & 0.3287 & 0.387 & 0.3033 \\ \hline
        Rideshare & 0.7485 & 0.7485 & 1.5189 & 1.9635 & N/A \\ \hline
        Temperature Rain & Timeout & 1.4579 & 0.966 & 1.6073 & N/A \\ \hline
        KDD Cup & 0.5138 & 0.4921 & 0.6273 & 0.5847 & N/A \\ \hline
    \end{tabular}
    \caption{Individual SMAPE scores}
    \label{table:4}
\end{table}

\begin{table}[!ht]
    \centering
    \begin{tabular}{ |p{2cm}|p{2.2cm}|p{2.2cm}|p{2.2cm}|p{2cm}|p{2cm}|  }
    % \hline
    % \multicolumn{6}{|c|}{SMAPE} \\
    % \hline
    \hline
        \textbf{Trend} & \textbf{Polynomial Trend} & \textbf{Auto ARIMA} & \textbf{Exponential Smoothing} & \textbf{AutoETS} & \textbf{Prophet} \\ \hline
         0.2133 & 0.2133 & 0.1724 & 0.2311 & 0.1917 & 0.1971 \\ \hline
        0.1822 & 0.1822 & 0.1843 & 0.1805 & 0.1826 & 0.1908 \\ \hline
        0.2175 & 0.2175 & 0.1674 & 0.1746 & 0.1818 & 0.2001 \\ \hline
        0.2292 & 0.2292 & 0.1706 & 0.1779 & 0.1619 & 0.2217 \\ \hline
        0.1445 & 0.1445 & 0.1185 & 0.1082 & 0.1104 & 0.1367 \\ \hline
        0.207 & 0.207 & 0.1737 & 0.1620 & 0.1658 & 0.2073 \\ \hline
        0.2183 & 0.2183 & Timeout & 0.1638 & 0.1397 & Timeout \\ \hline
        0.1883 & 0.1883 & Timeout & 0.1109 & 0.11 & Timeout \\ \hline
        0.2451 & 0.2451 & Timeout & 0.1437 & Timeout & Timeout \\ \hline
        0.2989 & 0.2989 & 0.1025 & 0.0901 & 0.0871 & 0.2021 \\ \hline
        0.201 & 0.201 & Timeout & 0.0304 & 0.0302 & Timeout \\ \hline
        0.3316 & 0.3316 & Timeout & 0.4295 & 0.4228 & 0.1808 \\ \hline
        0.2889 & 0.2889 & 0.3144 & 0.3554 & 0.348 & 0.2884 \\ \hline
        0.3183 & 0.3183 & 0.1664 & 0.2737 & 0.2822 & 0.2353 \\ \hline
        0.3695 & 0.3695 & 0.2127 & 0.3627 & 0.3739 & 0.2682 \\ \hline
        0.3594 & 0.3594 & 0.2244 & 0.3550 & 0.3561 & 0.2171 \\ \hline
        0.115 & 0.115 & 0.1312 & 0.1224 & 0.121 & 0.112 \\ \hline
        0.411 & 0.411 & 0.2563 & 0.2475 & 0.2036 & 0.4106 \\ \hline
        0.1501 & 0.1501 & 0.1515 & 0.1249 & 0.1265 & 0.148 \\ \hline
        0.1606 & 0.1606 & 0.1493 & 0.1440 & 0.1417 & 0.104 \\ \hline
        0.5278 & 0.5278 & 0.2312 & 0.2071 & 0.2067 & 0.4411 \\ \hline
        1.3374 & 1.3374 & Timeout & 1.2397 & 1.2404 & 1.2874 \\ \hline
        0.714 & 0.714 & Timeout & 0.7207 & 0.7235 & Timeout \\ \hline
        0.419 & 0.419 & Timeout & 0.4438 & 0.4441 & Timeout \\ \hline
        1.7977 & 1.7978 & N/A & 1.7606 & 1.7723 & 1.8119 \\ \hline
        0.2722 & 0.2722 & 0.1069 & 0.1065 & 0.1028 & 0.3026 \\ \hline
        0.1917 & 0.1917 & 0.2349 & 0.1803 & 0.1882 & 0.1935 \\ \hline
        0.4037 & 0.4037 & 0.0942 & 0.1622 & 0.0947 & 0.1768 \\ \hline
        1.9971 & 1.9971 & Timeout & 1.9620 & 1.9618 & 1.9918 \\ \hline
        0.1219 & 0.1219 & 0.0512 & 0.1180 & 0.1220 & 0.0623 \\ \hline
        0.3706 & 0.3706 & 0.3736 & 0.3603 & 0.3579 & 0.3795 \\ \hline
        N/A & N/A & N/A & N/A & N/A & N/A \\ \hline
        N/A & N/A & N/A & N/A & N/A & N/A \\ \hline
        1.3317 & 1.3317 & Timeout & 0.5239 & Timeout & Timeout \\ \hline
        0.4267 & 0.4267 & 0.3843 & 0.3686 & 0.3988 & 0.3670 \\ \hline
        2 & 2 & Timeout & 1.6525 & 1.6527 & 1.9985 \\ \hline
        1.464 & 1.464 & Timeout & 1.5118 & 1.5131 & Timeout \\ \hline
        0.6115 & 0.6115 & Timeout & 0.6220 & 0.6257 & 0.5071 \\ \hline
    \end{tabular}
    \caption{Individual SMAPE scores}
    \label{table:5}
\end{table}

\begin{table}[!ht]
    \centering
    \begin{tabular}{ |p{3.7cm}|p{1.8cm}|p{1.8cm}|p{1.8cm}|p{1.8cm}|p{1.8cm}|  }
    % \hline
    % \multicolumn{6}{|c|}{MASE} \\
    % \hline
    \hline
        \textbf{Dataset} & \textbf{AG-TS 600} & \textbf{AG-TS 3600} & \textbf{Naive} & \textbf{STL} & \textbf{Theta}\\ \hline
        M1 yearly & 3.5844 & 3.5561 & 4.8943 & N/A & 3.9924 \\ \hline
        M1 quarterly & 1.7528 & 1.7616 & 2.0775 & 1.7998 & 1.6471 \\ \hline
        M1 monthly & 1.196 & 1.1041 & 1.3145 & 1.1374 & 1.047 \\ \hline
        M3 yearly & 2.7133 & 2.6272 & 3.1717 & N/A & 2.7067 \\ \hline
        M3 quarterly & 1.3212 & 1.2219 & 1.4253 & 1.1556 & 1.104 \\ \hline
        M3 monthly & 1.1749 & 1.0086 & 1.1462 & 0.906 & 0.8834 \\ \hline
        M4 Yearly & 3.0194 & 3.0492 & 3.9744 & N/A & 3.2793 \\ \hline
        M4 Quarterly & 1.2287 & 1.1756 & 1.6022 & 1.2528 & 1.2198 \\ \hline
        M4 Monthly & Timeout & 1.0625 & 1.2597 & Timeout & 0.9609 \\ \hline
        M4 Weekly & 0.4909 & 0.4531 & 2.7773 & N/A & 2.6366 \\ \hline
        M4 Daily & 1.1253 & 1.1225 & 1.4518 & 1.3707 & 1.153 \\ \hline
        M4 Hourly & 1.4460 & 1.2830 & 1.1932 & 1.1621 & 2.4569 \\ \hline
        Tourism Yearly & 3.0984 & 2.9105 & 3.552 & N/A & 3.0637 \\ \hline
        Tourism Quarterly & 1.7926 & 1.6196 & 1.699 & 1.5034 & N/A \\ \hline
        Tourism Monthly & 1.7335 & 1.6421 & 1.6309 & 1.4676 & N/A \\ \hline
        NN5 Daily & 0.8609 & 0.8342 & 1.0113 & 1.0533 & N/A \\ \hline
        NN5 Weekly & 0.8529 & 0.8599 & 1.0628 & N/A & 0.9702 \\ \hline
        Solar Weekly & 768.3479 & 971.4901 & 1.735 & N/A & 1.2355 \\ \hline
        Traffic Weekly & 1.0963 & 1.1034 & 1.7743 & N/A & N/A \\ \hline
        Electricity Weekly & 0.854 & 0.9858 & 3.0372 & N/A & N/A \\ \hline
        Bitcoin Dataset & 4.7629 & 4.7629 & 6.1548 & 6.3612 & 5.0802 \\ \hline
        Melbourne Pedestrian Counts & 0.6477 & 0.387 & 0.3768 & 0.4553 & N/A \\ \hline
        Traffic Hourly & 1.7443 & 1.0163 & 0.9105 & 0.9575 & N/A \\ \hline
        Electricity Hourly & 1.6992 & 2.1576 & 2.1661 & 1.9407 & N/A \\ \hline
        Car Parts & 0.792 & 0.7472 & 2.20E+13 & 2.20E+13 & N/A \\ \hline
        FRED-MD & 0.5593 & 0.4463 & 1.1008 & 0.6966 & N/A \\ \hline
        Hospital & 0.9461 & 0.8126 & 0.9205 & 0.9082 & 0.7882 \\ \hline
        COVID-19 Deaths & 5.8893 & 4.844 & 3.07E+14 & 3.07E+14 & N/A \\ \hline
        Sunspot Daily  & 0.048 & 0.0033 & 0.4028 & 0.3204 & N/A \\ \hline
        US Births & 1.9597 & 1.4976 & 1.6998 & 1.9631 & 2.1367 \\ \hline
        Saugeen River Flow & 1.3272 & 1.3272 & 2.8965 & 2.4098 & 1.4259 \\ \hline
        Wind Power & 0.2473 & 0.2473 & N/A & N/A & N/A \\ \hline
        London Smart Meters & Timeout & 1.0239 & N/A & N/A & N/A \\ \hline
        Wind Farms Dataset & Timeout & 1.6470 & 1.2947 & N/A & N/A \\ \hline
        Vehicle Trips  & 2.2257 & 2.0615 & 1.37E+12 & 1.37E+12 & 1.37E+12 \\ \hline
        Rideshare & 2.1959 & 2.1959 & 4.0404 & 5.17 & N/A \\ \hline
        Temperature Rain & Timeout & 0.8336 & 1.4136 & 1.7212 & N/A \\ \hline
        KDD Cup & 1.2382 & 1.2515 & 1.6367 & 1.4828 & N/A \\ \hline
    \end{tabular}
    \caption{Individual MASE scores}
    \label{table:6}
\end{table}

\begin{table}[!ht]
    \centering
    \begin{tabular}{ |p{2cm}|p{2.2cm}|p{2cm}|p{2.2cm}|p{2cm}|p{2cm}|  }
    % \hline
    % \multicolumn{6}{|c|}{MASE} \\
    % \hline
    \hline
        \textbf{Trend} & \textbf{Polynomial Trend} & \textbf{Auto ARIMA} & \textbf{Exponential Smoothing} & \textbf{AutoETS} & \textbf{Prophet} \\ \hline
         4.4203 & 4.4203 & 3.4935 & 4.936 & 3.9502 & 3.9491 \\ \hline
        2.082 & 2.082 & 1.7545 & 1.9267 & 1.8549 & 2.1971 \\ \hline
        1.4781 & 1.4781 & 1.1959 & 1.381 & 1.4265 & 1.2555 \\ \hline
        3.8828 & 3.8828 & 2.9014 & 3.1709 & 2.6954 & 3.5777 \\ \hline
        1.7611 & 1.7611 & 1.3971 & 1.4126 & 1.3797 & 1.5196 \\ \hline
        1.2485 & 1.2485 & 0.9712 & 1.0902 & 1.0621 & 1.1093 \\ \hline
        5.0219 & 5.0219 & Timeout & 3.9821 & 3.0862 & Timeout \\ \hline
        2.4031 & 2.4031 & Timeout & 1.4168 & 1.2782 & Timeout \\ \hline
        1.929 & 1.929 & Timeout & 1.1497 & Timeout & Timeout \\ \hline
        21.8586 & 21.8586 & 2.5562 & 2.6845 & 2.5479 & 8.294 \\ \hline
        7.9282 & 7.9282 & Timeout & 1.1537 & 1.1392 & Timeout \\ \hline
        10.5521 & 10.5521 & Timeout & 11.6067 & 12.0429 & 1.6306 \\ \hline
        3.3023 & 3.3023 & 3.3136 & 3.3262 & 3.1832 & 3.2304 \\ \hline
        3.703 & 3.703 & 1.6664 & 3.2072 & 3.2796 & 2.5085 \\ \hline
        3.3554 & 3.3554 & 2.067 & 3.2966 & 3.6033 & 2.2228 \\ \hline
        1.5458 & 1.5458 & 0.8821 & 1.5211 & 1.5259 & 0.8528 \\ \hline
        0.935 & 0.935 & 0.9885 & 0.991 & 0.98 & 0.9099 \\ \hline
        2.2959 & 2.2959 & 1.2776 & 1.2245 & 0.9977 & 2.2933 \\ \hline
        2.0337 & 2.0337 & 1.8931 & 1.6829 & 1.7003 & 2.023 \\ \hline
        3.0134 & 3.0134 & 2.9291 & 3.0871 & 3.0764 & 1.8132 \\ \hline
        13.3123 & 13.3123 & 5.8081 & 5.1326 & 4.3446 & 10.9526 \\ \hline
        2.6417 & 2.6417 & Timeout & 0.9572 & 0.9575 & 2.0246 \\ \hline
        2.099 & 2.099 & Timeout & 1.9221 & 1.9221 & Timeout \\ \hline
        4.1209 & 4.1209 & Timeout & 4.543 & 4.5447 & Timeout \\ \hline
        2.20E+13 & 2.20E+13 & N/A & 2.20E+13 & 2.20E+13 & 2.20E+13 \\ \hline
        4.2876 & 4.2876 & 0.4972 & 0.6168 & 0.4463 & 2.3278 \\ \hline
        0.8609 & 0.8609 & 0.9054 & 0.8146 & 0.8585 & 0.8396 \\ \hline
        3.07E+14 & 3.07E+14 & 3.07E+14 & 3.07E+14 & 3.07E+14 & 3.07E+14 \\ \hline
        2.3097 & 2.3097 & Timeout & 0.1283 & 0.1273 & 0.8452 \\ \hline
        4.5032 & 4.5032 & 1.8977 & 4.3543 & 4.5072 & 2.2697 \\ \hline
        1.4692 & 1.4692 & 1.4815 & 1.4259 & 1.4161 & 1.4923 \\ \hline
        N/A & N/A & N/A & N/A & N/A & N/A \\ \hline
        N/A & N/A & N/A & N/A & N/A & N/A \\ \hline
        2.3709 & 2.3709 & Timeout & 1.5938 & Timeout & Timeout \\ \hline
        1.37E+12 & 1.37E+12 & 1.37E+12 & 1.37E+12 & 1.37E+12 & 1.37E+12 \\ \hline
        6.7054 & 6.7054 & Timeout & 4.0401 & 4.0402 & 6.5282 \\ \hline
        1.1115 & 1.1115 & Timeout & 1.3609 & 1.3941 & Timeout \\ \hline
        1.4167 & 1.4167 & Timeout & 1.6452 & 1.6731 & 1.2475 \\ \hline
    \end{tabular}
    \caption{Individual MASE scores}
    \label{table:7}
\end{table}

\begin{table}[!ht]
    \centering
    % \captionsetup{justification=centering, labelsep=period}
  %\begin{tabular}{@{}|l|cc|cc|@{}}
  \begin{tabular}{@{}|p{3.7cm}|p{2.2cm}| p{2.2cm}|p{2.2cm}|p{2.2cm}|@{}} 
        \hline
        \multirow{2}{*}{\textbf{Dataset}} & \multicolumn{2}{c|}{\textbf{SMAPE}} & \multicolumn{2}{c|}{\textbf{MASE}} \\
        \cline{2-5}
        & \textbf{Recursive Tabular} & \textbf{PatchTST} & \textbf{Recursive Tabular} & \textbf{PatchTST} \\ 
        \hline
        M1 yearly & 0.1664 & 0.2012 & 3.6487 & 4.1134 \\ \hline
        M1 quarterly & 0.1728 & 0.1697 & 1.8047 & 1.8741 \\ \hline
        M1 monthly & 0.2264 & 0.1621 & 1.4542 & 1.1271 \\ \hline
        M3 yearly & 0.1688 & 0.1680 & 2.6464 & 3.0400 \\ \hline
        M3 quarterly & 0.1115 & 0.1044 & 1.5571 & 1.3690 \\ \hline
        M3 monthly & 0.1886 & 0.1502 & 1.2096 & 1.0245 \\ \hline
        M4 Yearly & 0.1414 & 0.1621 & 3.0705 & 3.9312 \\ \hline
        M4 Quarterly & 0.1125 & 0.1080 & 1.2329 & 1.3744 \\ \hline
        M4 Monthly & Timeout & Timeout & Timeout & Timeout \\ \hline
        M4 Weekly & 0.0934 & 0.0706 & 2.3270 & 2.4065 \\ \hline
        M4 Daily & 0.0311 & 0.0353 & 1.1273 & 1.3131 \\ \hline
        M4 Hourly & 0.3578 & 0.1395 & 2.1292 & 1.4102 \\ \hline
        Tourism Yearly & 0.4352 & 0.3188 & 3.1854 & 3.4116 \\ \hline
        Tourism Quarterly & 0.2776 & 0.1623 & 1.8055 & 2.0825 \\ \hline
        Tourism Monthly & 0.3498 & 0.1977 & 2.5910 & 1.6591 \\ \hline
        NN5 Daily & 0.3965 & 0.2035 & 1.0570 & 0.6902 \\ \hline
        NN5 Weekly & 0.1232 & 0.1090 & 0.9791 & 0.7737 \\ \hline
        Solar Weekly & 0.3370 & 0.2960 & 1.7193 & 0.9531 \\ \hline
        Traffic Weekly & 0.1318 & 0.1324 & 1.7611 & 1.3145 \\ \hline
        Electricity Weekly & 0.1547 & 0.0916 & 3.0935 & 1.2994 \\ \hline
        Bitcoin & 0.3140 & 0.1983 & 5.1446 & 5.0444 \\ \hline
        Melbourne Pedestrian Counts & 0.8086 & 0.4938 & 0.6215 & 0.3728 \\ \hline
        Traffic Hourly & 0.8159 & 0.3908 & 11.5098 & 1.0702 \\ \hline
        Electricity Hourly & 0.4611 & 0.1957 & 4.5676 & 1.0924 \\ \hline
        Car Parts & 1.8554 & 1.9908 & 1.5055 & 0.7489 \\ \hline
        FRED-MD & 0.1059 & 0.1568 & 0.4706 & 0.8997 \\ \hline
        Hospital & 0.2246 & 0.1909 & 0.8580 & 0.9740 \\ \hline
        COVID-19 Deaths & 0.1862 & 0.3912 & 7.0335 & 7.2033 \\ \hline
        Sunspot Daily & 1.9624 & 1.9999 & 0.1140 & 0.0029 \\ \hline
        US Births & 0.0579 & 0.0441 & 1.8715 & 1.5590 \\ \hline
        Saugeen River Flow & 0.3424 & 0.3763 & 1.3001 & 1.5194 \\ \hline
        Wind Power & 0.4889 & 0.4774 & 0.3048 & 0.3808 \\ \hline
        London Smart Meters & Timeout & Timeout & Timeout & Timeout \\ \hline
        Wind Farms & Timeout & Timeout & Timeout & Timeout \\ \hline
        Vehicle Trips & 0.4490 & 0.2996 & 2.2502 & 1.9612 \\ \hline
        Rideshare & 1.9947 & 1.9845 & 4.4221 & 4.0179 \\ \hline
        Temperature Rain & Timeout & Timeout & Timeout & Timeout \\ \hline
        KDD Cup & 0.6148 & 0.4854 & 3.2662 & 0.8044 \\ \hline
    \end{tabular}
    \caption{\centering {Individual scores for RecursiveTabular and PatchTST trained with a time limit of 10 minutes}}
    \label{table:alone}
\end{table}

\begin{table}[!ht]
    \centering
    \begin{tabular}{|p{5cm}|p{2cm}|p{2cm}|p{2cm}|p{2cm}|}
    \hline
        \textbf{Methods} & \textbf{Wins} & \textbf{Losses} & \textbf{Tie} & \textbf{Failures} \\ \hline \hline
        Naive & 9 & 27 & 0 & 0 \\ \hline \hline
        STL & 2 & 25 & 0 & 10 \\ \hline \hline
        Theta & 7 & 13 & 0 & 16 \\ \hline \hline
        Trend & 0 & 36 & 0 & 0 \\ \hline \hline
        PolynomialTrend & 0 & 36 & 0 & 0 \\ \hline \hline
        StatsForecastAutoARIMA & 2 & 20 & 0 & 14 \\ \hline \hline
        ExponentialSmoothing & 2 & 34 & 0 & 0 \\ \hline \hline
        StatsForecastAutoETS & 10 & 24 & 0 & 2 \\ \hline \hline
        Prophet & 5 & 24 & 0 & 8 \\ \hline 
    \end{tabular}
    \caption{Wins and losses table for SMAPE}
    \label{table:1}
\end{table}

\vskip 0.2in

\begin{table}[!ht]
    \centering
    \begin{tabular}{|p{5cm}|p{2cm}|p{2cm}|p{2cm}|p{2cm}|}
    \hline
        \textbf{Methods} & \textbf{Wins} & \textbf{Losses} & \textbf{Tie} & \textbf{Failures} \\ \hline \hline
        Naive & 4 & 32 & 0 & 0 \\ \hline \hline
        STL & 4 & 22 & 0 & 10 \\ \hline \hline
        Theta & 8 & 12 & 0 & 16 \\ \hline \hline
        Trend & 0 & 35 & 1 & 0 \\ \hline \hline
        PolynomialTrend & 0 & 35 & 1 & 0 \\ \hline \hline
        StatsForecastAutoARIMA & 1 & 21 & 0 & 14 \\ \hline \hline
        ExponentialSmoothing & 2 & 34 & 0 & 0 \\ \hline \hline
        StatsForecastAutoETS & 9 & 25 & 0 & 2 \\ \hline \hline
        Prophet & 4 & 24 & 0 & 8 \\ \hline 
    \end{tabular}
    \caption{Wins and losses table for MASE}
    \label{table:2}
\end{table}

\begin{table}[!ht]
    \centering
    \begin{tabular}
    {|| p{2.4cm}|p{1.7cm}|p{1.7cm}|p{1.7cm}|p{1.7cm}|p{2cm}|p{2.2cm} ||}
    \hline \hline
        \textbf{Dataset Name} & \textbf{Naive} & \textbf{STL} & \textbf{Theta} & \textbf{Trend} & \textbf{Polynomial Trend} & \textbf{Exponential Smoothing} \\ \hline \hline
        M1 yearly & 0.2243 & N/A & 0.1922 & 0.2441 & 0.2437 & 0.4409 \\ \hline
        M1 quarterly & 0.1735 & N/A & 0.167 & 0.1822 & 0.1812 & 0.1891 \\ \hline
        M1 monthly & 0.1731 & 0.1867 & 0.1927 & 0.1781 & 0.1795 & 0.1924 \\ \hline
        M3 yearly & 0.1788 & N/A & 0.1673 & 0.1917 & 0.2292 & 0.2392 \\ \hline
        M3 quarterly & 0.1107 & 0.0997 & 0.0914 & 0.1091 & 0.1095 & 0.1279 \\ \hline
        M3 monthly & 0.1724 & 0.166 & 0.1442 & 0.1727 & 0.1727 & 0.164 \\ \hline
        M4 monthly & Timeout & Timeout & Timeout & Timeout & Timeout & Timeout \\ \hline
        M4 Quarterly & 0.1252 & Timeout & 0.1034 & Timeout & Timeout & 0.1262 \\ \hline
        M4 Weekly & 0.0916 & N/A & 0.0911 & 0.0911 & 0.0905 & 0.0997 \\ \hline
        M4 Hourly & Timeout & Timeout & Timeout & Timeout & Timeout & 0.3517 \\ \hline
        Tourism Yearly & 0.4217 & N/A & 0.3343 & 0.289 & 0.3237 & 0.358 \\ \hline
        Tourism Quarterly & 0.6066 & 0.1576 & 0.2693 & 0.2766 & 0.2764 & 0.3102 \\ \hline
        Tourism Monthly & 0.2167 & 0.1961 & 0.3574 & 0.3762 & 0.377 & 0.3292 \\ \hline
        NN5 Daily & 0.2275 & 0.2447 & 0.3558 & 0.4083 & Timeout & 0.3614 \\ \hline
        NN5 Weekly & 0.1435 & N/A & 0.1203 & 0.1255 & 0.124 & 0.1211 \\ \hline
        Solar Weekly & 0.328 & N/A & 0.2528 & 0.2728 & 0.273 & 0.2486 \\ \hline
        Traffic Weekly & 0.1307 & N/A & 0.1258 & 0.1251 & 0.1615 & 0.1276 \\ \hline
        Bitcoin  & 1.3922 & Timeout & 0.2112 & Timeout & 0.2527 & 0.2127 \\ \hline
        Car Parts & 0.6543 & 1.4473 & 1.7886 & 1.0948 & 1.6394 & 1.7872 \\ \hline
        FRED-MD & 0.1069 & 0.1223 & 0.1069 & 0.1079 & Timeout & 0.1145 \\ \hline
        Hospital & 0.2125 & 0.1925 & 0.181 & 0.1885 & 0.1939 & 0.1805 \\ \hline
        COVID19 Deaths & 0.8646 & 0.1993 & 0.1637 & 0.4037 & 0.3028 & 0.2377 \\ \hline
        US Births & 0.045 & 0.0412 & 0.0582 & 0.1136 & 0.1121 & 0.1191 \\ \hline
        Saugeen River Flow & 0.3603 & Timeout & 0.3603 & Timeout & Timeout & 0.4557 \\ \hline
        Vehicle Trips & 0.3287 & 0.3421 & 0.3813 & 0.3657 & 0.3649 & 0.3634 \\ \hline
        Rideshare & Timeout & Timeout & 2 & Timeout & Timeout & 1.8749 \\ \hline 
    \end{tabular}
    
    \caption{Individual SMAPE scores after Tuning}
    \label{smape_tuning}
\end{table}

\newpage
\begin{table}[!ht]
    \centering
    \begin{tabular}{||p{2.4cm}|p{1.7cm}|p{1.7cm}|p{1.7cm}|p{1.7cm}|p{2cm}|p{2.2cm}||}
    \hline \hline
        \textbf{Dataset Name} & \textbf{Naive} & \textbf{STL} & \textbf{Theta} & \textbf{Trend} & \textbf{Polynomial Trend} & \textbf{Exponential Smoothing} \\ \hline \hline
        M1 yearly & 4.8943 & N/A & 3.9924 & 5.2791 & 5.2851 & 8.7938 \\ \hline
        M1 quaterly & 1.8095 & N/A & 1.7638 & 1.9784 & 1.9786 & 2.127 \\ \hline
        M1 monthly & 1.3145 & 1.1989 & 1.3571 & 1.3953 & 1.4033 & 1.4932 \\ \hline
        M3 yearly & 3.1717 & N/A & 2.7067 & 3.4618 & 3.8828 & 4.435 \\ \hline
        M3 quaterly & 1.4253 & 1.155 & 1.104 & 1.4316 & 1.4337 & 1.7738 \\ \hline
        M3 monthly & 1.1462 & 1.1129 & 0.8834 & 1.1369 & 1.1378 & 1.1097 \\ \hline
        M4 monthly & Timeout & Timeout & Timeout & Timeout & Timeout & Timeout \\ \hline
        M4 Quarterly & 1.6022 & Timeout & 1.2198 & Timeout & Timeout & 1.6818 \\ \hline
        M4 Weekly & 2.7773 & N/A & 2.6366 & 2.7285 & 2.724 & 3.0399 \\ \hline
        M4 Hourly & Timeout & Timeout & Timeout & Timeout & Timeout & 10.6989\\ \hline
        Tourism Yearly & 3.552 & N/A & 3.0637 & 3.3038 & 3.2588 & 4.3058 \\ \hline
        Tourism Quarterly & 6.1699 & 1.6354 & 3.1542 & 3.2651 & 3.2617 & 3.6157 \\ \hline
        Tourism Monthly & 1.6309 & 1.4918 & 3.2517 & 3.3963 & 3.4034 & 3.003 \\ \hline
        NN5 Daily & 0.9342 & 0.9557 & 1.5252 & 1.6722 & Timeout & 1.5439 \\ \hline
        NN5 Weekly & 1.205 & N/A & 0.9702 & 1.0084 & 0.9974 & 0.9807 \\ \hline
        Solar Weekly & 1.735 & N/A & 1.2556 & 1.3809 & 1.3822 & 1.2294 \\ \hline
        Traffic Weekly & 1.7743 & N/A & 1.6988 & 1.6937 & 2.1907 & 1.7099 \\ \hline
        Bitcoin & 29.4626 & Timeout & 5.6416 & Timeout & 6.4721 & 5.6611 \\ \hline
        Car Parts & 2.2E+13 & 2.2E+13 & 2.2E+13 & 2.2E+13 & 2.2E+13 & 2.2E+13 \\ \hline
        FRED-MD & 0.5657 & 0.7863 & 0.5775 & 0.6492 & Timeout & 0.774 \\ \hline
        Hospital & 0.97 & 0.8366 & 0.8184 & 0.8591 & 0.8703 & 0.8078 \\ \hline
        COVID-19 Deaths & 3.07E+14 & 3.07E+14 & 3.07E+14 & 3.07E+14 & 3.07E+14 & 3.07E+14 \\ \hline
        US Births & 1.6998 & 1.5496 & 2.1362 & 4.2027 & 4.1397 & 4.3972 \\ \hline
        Saugeen River Flow & 1.4257 & N/A & 1.4259 & Timeout & Timeout & 1.7139 \\ \hline
        Vehicle Trips & 1.37E+12 & 1.37E+12 & 1.37E+12 & 1.37E+12 & 1.37E+12 & 1.37E+12 \\ \hline
        Rideshare & Timeout & Timeout & 4.8814 & Timeout & Timeout & 4.0404 \\ \hline 
    \end{tabular}
    \caption{Individual MASE scores after Tuning}
    \label{mase_tuning}
\end{table}

\clearpage

\begin{table}[!ht]
    \centering
    \begin{tabular}{ |p{3.7cm}|p{2cm}|p{2cm}|p{2cm}|p{1.8cm}|p{2cm}|  }
    \hline
        \textbf{Dataset} & \textbf{AG-TS 600} & \textbf{AG-TS 3600} & \textbf{Naive} & \textbf{STL} & \textbf{Theta}\\ \hline
        M1 yearly & 6.68E+05 & 7.46E+05 & 195906.1904 & N/A & 167174.6985 \\ \hline
        M1 quaterly & 6919.0604 & 6053.7638 & 3234.9771 & 1974.6565 & 2095.7041 \\ \hline
        M1 monthly & 20286.1047 & 20286.1048 & 2458.1691 & 2609.4445 & 2374.1004 \\ \hline
        M3 yearly & 1594.7078 & 1532.9630 & 1178.5891 & N/A & 1100.7389 \\ \hline
        M3 quarterly & 973.5916 & 936.6338 & 682.2061 & 600.9747 & 564.0433 \\ \hline
        M3 monthly & 1373.3787 & 1373.6694 & 951.0313 & 847.6004 & 802.4535 \\ \hline
        M4 Yearly & 1834.9305 & 1779.3635 & 1153.1219 & N/A & 1001.8283 \\ \hline
        M4 Quarterly & 1464.3852 & 1344.5914 & 822.8239 & 721.6619 & 675.0591 \\ \hline
        M4 Monthly & Timeout & 1417.4375 & 846.7097 & Timeout & 679.6451 \\ \hline
        M4 Weekly & 604.0349 & 547.9879 & 423.8021 & N/A & 412.9419 \\ \hline
        M4 Daily & 675.0083 & 627.1100 & 262.1636 & 245.7951 & 210.2485 \\ \hline
        M4 Hourly & 1153.7589 & 1221.3757 & 426.3349 & 436.509 & 478.2892 \\ \hline
        Tourism Yearly & 3.70E+05 & 4.00E+05 & 111078.1247 & N/A & 104455.4863 \\ \hline
        Tourism Quarterly & 48560.8943 & 48279.9276 & 14072.4090 & 11900.9340 & N/A \\ \hline
        Tourism Monthly & 12501.3218 & 9566.0208 & 2575.6646 & 2367.9114 & N/A \\ \hline
        NN5 Daily & 5.6603 & 5.6957 & 6.1925 & 6.2187 & N/A \\ \hline
        NN5 Weekly & 20.0725 & 20.6291 & 20.2073 & N/A & 18.608 \\ \hline
        Solar Weekly & 1017.0866 & 981.3025 & 1918.4411 & N/A & 1353.6958 \\ \hline
        Traffic Weekly & 1.9454 & 1.9098 & 1.5367 & N/A & N/A \\ \hline
        Electricity Weekly & 2.75E+05 & 2.52E+05 & 75198.8219 & N/A & N/A \\ \hline
        Bitcoin Dataset & 3.83E+18 & 3.78E+18 & 1.96E+18 & 1.99E+18 & 1.55E+18 \\ \hline
        Melbourne Pedestrian Counts & 204.6244 & 244.1703 & 94.2956 & 95.1918 & N/A \\ \hline
        Traffic Hourly & 0.0379 & 0.0266 & 0.0221 & 0.0216 & N/A \\ \hline
        Electricity Hourly & 1717.8993 & 1721.0808 & 606.7372 & 531.5058 & N/A \\ \hline
        Car Parts & 1.1184 & 1.0716 & 1.0628 & 1.1500 & N/A \\ \hline
        FRED-MD & 27824.0869 & 7782.7724 & 5578.1401 & 4147.6196 & N/A \\ \hline
        Hospital & 75.4072 & 69.5344 & 25.3392 & 24.4388 & 21.6453 \\ \hline
        COVID-19 Deaths & 430.6109 & 430.6109 & 460.8636 & 312.9606 & N/A \\ \hline
        Sunspot Daily  & 1.0138 & 0.7718 & 17.6446 & 12.5311 & N/A \\ \hline
        US Births & 690.1375 & 579.735 & 711.6357 & 767.1003 & 735.3762 \\ \hline
        Saugeen River Flow & 38.03822 & 38.0382 & 51.0880 & 43.1261 & 39.7721 \\ \hline
        Wind Power & 13.1803 & 13.1803 & N/A & N/A & N/A \\ \hline
        London Smart Meters  & Timeout & 0.2655 & N/A & N/A & N/A \\ \hline
        Wind Farms  & Timeout & 54.1845 & 22.3619 & N/A & N/A \\ \hline
        Vehicle Trips  & 67.2058 & 62.8733 & 29.5722 & 30.6123 & 27.8123 \\ \hline
        Rideshare  & 6.0708 & 6.1376 & 7.1745 & 8.7476 & N/A \\ \hline
        Temperature Rain  & Timeout & 12.2862 & 12.9093 & 12.4037 & N/A \\ \hline
        KDD Cup & 136.0083 & 137.919 & 72.7309 & 69.0548 & N/A \\ \hline
    \end{tabular}
    \caption{Individual RMSE scores}
    \label{table:8}
\end{table}

\begin{table}[!ht]
    \centering
    \begin{tabular}{ |p{2.2cm}|p{2.2cm}|p{2.2cm}|p{2.2cm}|p{2.2cm}|p{2.2cm}|  }
    \hline
        \textbf{Trend} & \textbf{Polynomial Trend} & \textbf{Auto ARIMA} & \textbf{Exponential Smoothing} & \textbf{Auto ETS} & \textbf{Prophet} \\ \hline
         188635.6474 & 188635.6474 & 153235.4167 & 193902.9209 & 158815.7943 & 143408.4939 \\ \hline
        2264.8858 & 2264.8858 & 2483.9465 & 2543.7867 & 2413.8612 & 1876.7357 \\ \hline
        3105.8956 & 3105.8956 & 2666.7876 & 2739.7400 & 3025.6718 & 2838.3901 \\ \hline
        1463.4362 & 1463.4362 & 1322.0455 & 1173.5990 & 1127.9498 & 1425.8074 \\ \hline
        818.7577 & 818.7577 & 687.4503 & 668.3678 & 684.2575 & 778.5016 \\ \hline
        981.0756 & 981.0756 & 847.4839 & 891.6311 & 902.0841 & 948.0332 \\ \hline
        1376.3732 & 1376.3732 & Timeout & 1153.8372 & 961.1576 & Timeout \\ \hline
        1118.4622 & 1118.4622 & Timeout & 732.8815 & 716.2729 & Timeout \\ \hline
        1074.3009 & 1074.3009 & Timeout & 755.1853 & Timeout & Timeout \\ \hline
        1539.6767 & 1539.6767 & 424.2115 & 412.4430 & 405.4063 & 776.9076 \\ \hline
        1243.5188 & 1243.5188 & Timeout & 209.7506 & 208.2704 & Timeout \\ \hline
        1310.5191 & 1310.5191 & Timeout & 1477.4988 & 1478.548169 & 450.413 \\ \hline
        107231.017 & 107231.017 & 80952.2459 & 111515.6365 & 96626.6635 & 112894.653 \\ \hline
        21825.5539 & 21825.5539 & 13907.8186 & 17239.9344 & 18025.4992 & 12309.2721 \\ \hline
        7218.6665 & 7218.6665 & 2734.5887 & 7032.1298 & 7596.2541 & 3803.6284 \\ \hline
        8.2053 & 8.2053 & 5.3881 & 8.2321 & 8.2307 & 5.1333 \\ \hline
        18.4162 & 18.4162 & 18.8333 & 18.8111 & 18.5297 & 18.1178 \\ \hline
        2392.7561 & 2392.7561 & 1404.5680 & 1341.5244 & 1109.3207 & 2390.5520 \\ \hline
        1.7526 & 1.7526 & 1.6652 & 1.5141 & 1.5289 & 1.7407 \\ \hline
        80944.2452 & 80944.2452 & 72698.7224 & 76993.4413 & 72811.5137 & 25680.9758 \\ \hline
        9.02E+17 & 9.02E+17 & 1.74E+18 & 1.52E+18 & 1.21E+18 & 4.09E+18 \\ \hline
        482.2367 & 482.2367 & Timeout & 228.0643 & 228.1535 & 412.8096 \\ \hline
        0.0366 & 0.0366 & Timeout & 0.0351 & 0.03505 & Timeout \\ \hline
        1008.0578 & 1008.0578 & Timeout & 1025.6299 & 1026.2815 & Timeout \\ \hline
        0.8409 & 0.8409 & N/A & 0.7670 & 0.8031 & 1.01920 \\ \hline
        10053.778 & 10053.778 & 2411.9063 & 3104.7099 & 2216.0638 & 14841.5059 \\ \hline
        28.7643 & 28.7643 & 24.8585 & 26.5455 & 27.3348 & 24.897 \\ \hline
        816.0976 & 816.0976 & 96.6420 & 403.4973 & 106.7755 & 175.0196 \\ \hline
        89.0617 & 89.0617 & Timeout & 4.9609 & 4.9208 & 32.6130 \\ \hline
        1361.5918 & 1361.5918 & 700.6916 & 1368.2501 & 1360.5510 & 780.3741 \\ \hline
        43.1209 & 43.1209 & 43.8347 & 39.794 & 39.3295 & 43.6398 \\ \hline
        N/A & N/A & N/A & N/A & N/A & N/A \\ \hline
        N/A & N/A & N/A & N/A & N/A & N/A \\ \hline
        30.5140 & 30.5140 & Timeout & 26.9630 & Timeout & Timeout \\ \hline
        38.6727 & 38.6727 & 26.4495 & 36.6198 & 39.0391 & 30.0111 \\ \hline
        11.4951 & 11.4951 & Timeout & 7.175 & 7.1745 & 10.7885 \\ \hline
        8.7676 & 8.7676 & Timeout & 10.3412 & 10.3772 & Timeout \\ \hline
        73.9219 & 73.9219 & Timeout & 73.8121 & 74.2149 & 67.181 \\ \hline
    \end{tabular}
    \caption{Individual RMSE scores}
    \label{table:9}
\end{table}

\clearpage
\bibliography{sample}

\end{document}